\documentclass[review]{elsarticle}

\usepackage{lineno,hyperref}
\modulolinenumbers[5]

\journal{arXiv}

\usepackage{graphicx}
\usepackage[cmex10]{amsmath}
\usepackage{algorithmic}
\usepackage{array}
\usepackage{mdwmath}
\usepackage{mdwtab}
\usepackage{eqparbox}
\usepackage{subfig}
\usepackage{fixltx2e}
\usepackage{stfloats}
\usepackage{url}
\usepackage{amsfonts,amssymb}
\usepackage{pifont}
\usepackage{ctable}
\usepackage{multirow}
\usepackage{diagbox}

\usepackage{algorithmic}
\usepackage[ruled]{algorithm}

\newcommand{\tabincell}[2]{\begin{tabular}{@{}#1@{}}#2\end{tabular}}
\newcommand{\eat}[1]

\newcommand{\matrx}[1]{\boldsymbol{\rm #1}}
\newcommand{\vect}[1]{\boldsymbol{\rm #1}}

\graphicspath{{eps/}}

\begin{document}
	
\begin{frontmatter}
	
	\title{Classification of Neurological Gait Disorders Using Multi-task Feature Learning}

	\author[UCONN]{Ioannis Papavasileiou \footnote{The first two authors contributed equally to this work.}}
	\author[ASU]{Wenlong Zhang}
	\author[PHILIPS]{Xin Wang \footnote{Xin Wang was with Department of Computer Science \& Engineering at University of Connecticut.}}
	\author[UCONN]{Jinbo Bi}
	\author[NANJ]{Li Zhang}
	\author[UCONN]{Song Han}
		
	\address[UCONN]{ Department of Computer Science \&
		Engineering, University of Connecticut, Storrs, CT 06269 E-mails:
		$\{ioannis.papavasileiou, jinbo.bi, song.han\}@uconn.edu$}
	\address[ASU]{The Polytechnic School, Ira A. Fulton Schools of Engineering, Arizona State University, Mesa, AZ 85212 E-mail: $wenlong.zhang@asu.edu$}
	\address[PHILIPS]{Dept. of Clinical Informatics Solutions and Services, Philips Research North America E-mail: $xin.wang@philips.com$}
	\address[NANJ]{Dept. of Geriatric Neurology, Nanjing Brain Hospital, P. R. China E-mail: $jmuzhangli@sina.com$}
\begin{abstract}
As our population ages, neurological impairments and degeneration of
the musculoskeletal system yield gait abnormalities, which can significantly
reduce quality of life. Gait rehabilitative therapy has been widely
adopted to help these patients maximize community participation and living independence.
To further improve the precision and efficiency of rehabilitative therapy, more objective methods need to be developed based on sensory data.
In this paper, an algorithmic framework is proposed to provide classification of gait disorders caused by two common neurological diseases, stroke and Parkinson's Disease (PD), from ground contact force (GCF) data.
An advanced machine learning method, multi-task feature learning (MTFL), is used to jointly train classification models of a subject's gait in three classes, post-stroke, PD and healthy gait.
Gait parameters related to mobility, balance, strength and rhythm are used as features for the classification.
Out of all the features used, the MTFL models capture the more important ones per disease, which will help provide better objective assessment and therapy progress tracking.
To evaluate the proposed methodology we use data from a human participant study, which includes five PD patients, three post-stroke patients, and three healthy subjects. Despite the diversity of abnormalities, the evaluation shows that the proposed approach can successfully distinguish post-stroke and PD gait from healthy gait, as well as post-stroke from PD gait, with Area Under the Curve (AUC) score of at least 0.96. Moreover, the methodology helps select important gait features to better understand the key characteristics that distinguish abnormal gaits and design personalized treatment.
\end{abstract}

\end{frontmatter}


\section{Introduction}
Aging is an unprecedented, pervasive, profound and enduring process for humanity, and it is currently a global phenomenon~\cite{JRNL:AgingReport}. One major challenge associated with aging is the
degenerative conditions of the neuromusculoskeletal system (e.g.
osteoporosis, arthritis, Alzheimer's disease~\cite{MISC:alz},
stroke~\cite{MISC:stroke}, and Parkinson's disease~\cite{MISC:pd}). Any dysfunction of the central nervous system, spinal cord, peripheral nerves or muscles can result in an abnormal gait~\cite{BOOK:delisa1998gait}. At the age of 60, 85\% of people have a normal gait, while at the age of 85 or older this proportion drops to 18\%~\cite{snijders2007neurological}.
As a result, an increasing number of people suffer from walking
difficulties, and the demand for gait rehabilitative therapy has been increasing rapidly.

In the current practice, gait rehabilitative therapy is provided by therapists who manually stimulate patients' reflexes and rotate their lower limbs to retrain their central nervous systems with the correct gait patterns.
This approach is not only physically demanding for both patients and
therapists, but also expensive and time-consuming. Moreover, in the
clinic, assessment of gait abnormalities is based on timed tests,
visual observations by therapists, retrospective qualitative
evaluations of video tapes, and specific physical tests, e.g., strength,
range of motion, balance, gait speed, and endurance.
As a result, most times gait assessment is based on the subjective judgment of the therapist. More objective methods are desired to quantify the gait assessment and progress evaluation of the rehabilitative training, reduce the chances of biased assessment by therapists, and provide better, targeted treatment to patients.

Significant research efforts have been reported to provide more objective gait assessment.
A variety of sensory devices have been employed for gait analysis and impairment diagnosis. For instance, encoders, inertial sensors, and camera-based motion capture systems have been employed for kinematic analysis of human motion~\cite{jrnl:IMUrehab,jrnl:camerarehab}; force sensors~\cite{JRNL:forceplate,jrnl:mitss} and electromyography (EMG) sensors~\cite{jrnl:EMGrehab} have been widely used to study the ground contact forces (GCFs) and muscle activities during walking; electroencephalography (EEG) sensors have been employed to analyze brain signals~\cite{JRNL:gwin2010removal,JRNL:sburlea2015detecting} and better understand neurological mechanisms of walking.
Advanced signal processing methods have been designed and applied on the data recorded from such sensor platforms~\cite{JRNL:sant2011new,JRNL:tahir2012parkinson,JRNL:lee2012parkinson,mulroy2003use,kinsella2008gait,JRNL:chen2005gait}.
These sensor technologies can be used not only to detect or prognose various human gait disorders, but also for disease monitoring and therapy progress tracking and evaluation~\cite{snijders2007neurological}.

To better quantify the severity of abnormal gait, important sensing features need to be identified from the sensory data to characterize gait disorders. Towards this goal, extensive research efforts have been reported to use machine learning algorithms for gait classification and clustering, to identify such parameters and automate gait disorder diagnosis. For example, post-stroke patients usually experience a very diverse set of gait abnormalities, most common of which is the hemiplegic gait~\cite{BOOK:delisa1998gait}. For this reason, researchers have applied cluster analysis to identify subgroups of patients with similar sensing features who experience similar gait abnormalities \cite{mulroy2003use, kinsella2008gait, JRNL:chen2005gait}. Likewise, other research efforts focus on classifying abnormal gaits between healthy subjects and Parkinson's disease (PD) patients \cite{JRNL:tahir2012parkinson, JRNL:lee2012parkinson, JRNL:wu2010statistical}. Classification methods with feature selection can help the target design of treatment and evaluation of therapy through the identified important gait sensing features~\cite{chantraine2016proposition}. Furthermore, such tools can improve the valuable clinical management of the patients, ease communication between clinicians \cite{chantraine2016proposition} and optimize subject selection for human participant studies~\cite{ferrante2011biofeedback}. Consequently, they reduce the cost of physical therapy and improve the quality of life for patients. Especially, patients living in remote areas can benefit from an enhanced tele-medicine system with these quantitative and diagnostic tools, without necessitating complex apparatus \cite{chantraine2016proposition}. However, to the best of our knowledge, there is no such quantitative gait diagnostic system for neurological diseases.

In order to bridge this gap and enable objective gait analysis, we propose an integrative framework in this paper to automatically classify gait disorders from two common neurological diseases, stroke and PD, and distinguish abnormal gait caused by these two diseases from the healthy gait.
Classifying gait into groups caused by these two major neurological diseases can lead the way to provide diagnostic tools for specific gait disorders caused by these two neurological diseases, which is much needed for assisting objective gait assessment in the clinic and rehabilitation therapy centers.
Our integrative framework includes a pair of smart shoes as the sensory device to capture the GCF data and a pipeline of data analytic algorithms for feature extraction, classification and feature selection. Gait features, including mobility, balance, strength and rhythm, are extracted from the sensory data.

Because there is strong correlation between the two neurological diseases and resultant gait disorders, multi-task machine learning strategies can be more feasible to identify similarities and differences of gait patterns than classic multi-class classification algorithms given the latter focus on modeling only the exclusive (or discriminative) features of different gait classes~\cite{JMLR:v17:15-234,obozinski2006multi}. An advanced multi-task learning algorithm has been developed to jointly create three classifiers, respectively, for distinguishing stroke-induced gait from healthy gait, PD-induced gait from healthy gait, and PD-induced gait from stroke-induced gait. To evaluate the proposed methodology we use data from a human participant study, which includes five PD patients, three post-stroke patients, and three healthy subjects. In our experiments the classification performance achieved Area Under the Curve (AUC) score of at least 0.96.
The advantage of our multi-task learning method is that it can identify features useful for all three classification tasks as well as those predictive of a specific abnormality. We conclude our evaluation with a discussion on the important sensing features identified by the algorithms.

\eat{
	(In this paper we present a neurological gait disorder diagnosis algorithmic framework for classifying gait affected by two common neurological diseases, stroke and PD from healthy gait, using ground contact force (GCF) data. To our knowledge, there is no other research efforts that provide such gait disorder diagnosis framework for these two common neurological diseases. The framework is tested on an analytic computing platform which includes a pair of smart shoes
	as the sensory device to capture the GCF data and an analytics platform for efficient data storage and processing of the proposed analytic methods.
	Gait parameters regarding mobility, balance, strength and gait phases are used to capture
	the key characteristics of the underlying gait disorders that stroke and PD patients experience. Multi-task learning (MTL), an advanced classification method, is used to jointly build three classification tasks, which can distinguish between stroke, Parkinson's and healthy gait. The tasks are healthy vs stroke, healthy vs PD and stroke vs PD patient's gait. Overall and per task important gait parameters are identified. Data from a clinical study are used to evaluate this framework including five PD patients, three post-stroke patients and three healthy subjects.)}

The remainder of the paper is organized as follows. Section \ref{sec:related} reviews the related works in gait quantification and analysis. Section  \ref{sec:smartShoes} presents the sensory device that we developed to measure the GCF and Section \ref{sec:gaitParams} discusses the gait sensing features we extracted based on the data. In Section \ref{sec:mtl}, we introduce the multi-task learning approach and use it to classify gait based on the extracted sensing features. Evaluation results are given based on the recorded data from a human participant study and findings are summarized in Section \ref{sec:expEval}. We conclude the paper and discuss future work in Section \ref{sec:conclusion}.

Extensive research efforts have been made towards quantitative gait analysis.
In this section, we first discuss the literature studies  on improving gait quantification methods for objective gait parameter extraction. We then present a summary on machine learning methods for improving gait analysis, which includes gait pattern classification and cluster analysis for finding subgroups of patients who suffer from the same neurological disease and experience similar gait abnormalities.

\eat{
	\subsection{Smart rehabilitation}
	There is growing interest in utilizing smart sensing devices to facilitate gait rehabilitation in ambulatory settings. This is based on the observation that traditional methods used by medical professionals are not always trustful. For example, in \cite{lemoyne2008accelerometers} conflicting opinions are expressed on the reliability of qualitative ordinal scales used for quantifying movement. These methods rely on the objective judgment of the medical professionals and inherently do not consider temporal parameters. Traditional devices \cite{lemoyne2008accelerometers} used for objective gait analysis in an ambulatory setting include foot stride and GCF analyzers, electrogoniometers, electromyography (EMG), metabolic energy expenditure and optic sensors \cite{lee2014concurrent}. Smart and wearable sensing devices, interactive systems and robotics are going to be discussed next.
	
	Wearable smart devices offer an inexpensive, convenient and efficient way to provide useful information for gait analysis. The gait analysis methods based on wearable sensors can be generally divided into gait kinematics, gait kinetics and EMG~\cite{tao2012gait}. Gait kinematics must be established on the basis of kinematic measurement and analysis and the type of sensors used for this purpose are accelerometers, gyroscopes, magnetoresistive sensors, flexible goniometers, and sesning fabric, etc. Gait kinetics is the study of forces and moments that result in the movement of body segments in a human gait, including the measurement of GCF and kinetic analysis \cite{tao2012gait,rueterbories2010methods}. Kinematics can be measured by placing force plates on the ground, but several recent research works have developed wearable sensors and deployed them in the shoe for the purpose to measure GCF. In \cite{bamberg2008gait} a smart shoe system has been introduced to sense different modalities. The sensor suite includes three orthogonal accelerometers, three orthogonal gyroscopes, four force sensors, two bidirectional bend sensors, two dynamic pressure sensors, as well as electric field height sensors. In addition to that, sensors can be placed in various places on the trunk \cite{hartmann2009reproducibility}, waist \cite{yang2011real}, joints and lower extremity including feet \cite{mariani2013shoe}, and ankle \cite{lemoyne2009wireless}. Other smart non-wearable systems include vision sensors, like light beam and infrared thermography, and ultrasonic wave sensors \cite{muro2014gait}.
	
	The use of off-the-shelf interactive gaming systems is very attractive in the context of implementing home interventions, but commercially available systems lack the ability of monitoring movement patterns in a way that is satisfactory from a rehabilitation intervention standpoint \cite{bonato2010wearable}. In \cite{stone2011passive} Microsoft Kinect and web cameras were used to analyze the stride-to-stride variability passively in a home setting. In \cite{gabel2012full} a single Kinect sensor was used at home setting along with the virtual skeleton model provided for accurate measurements of a rich set of gait features.
	
	In addition to the smart sensing and interactive technologies, many merging interactive and robot technologies have paved the way towards improved home interventions and higher reliability in the detection of emergency situation \cite{bonato2010wearable}. Robotic training for the lower extremity coupled with a Virtual Reality system for feedback has been used for gait rehabilitative trainings of post-stroke patients \cite{mirelman2009effects, deutsch2004development}. In~\cite{belda2011rehabilitation} they examine possible benefits of including assistive robotic devices and brain-computer interfaces (BCI) in this field, according to a top-down approach, in which rehabilitation is driven by neural plasticity.
}

\subsection{Gait quantification}
Gait quantification is important for objective gait assessment and analysis. It relates to the methods used for objectively measuring gait parameters, which can be used to estimate the severity of human gait abnormality. In this subsection we discuss gait quantification with respect to hemiplegic and Parkinsonian gait, which are the two most popular gait disorders caused by stroke and PD respectively \cite{BOOK:delisa1998gait}.

Among many gait parameters, symmetry is an important gait characteristic and is defined as a perfect agreement between the actions of the two lower limbs \cite{sadeghi2000symmetry}. To calculate symmetry, mobility parameters (e.g., single support ratio) and spatiotemporal parameters (e.g., step length) can be used~\cite{patterson2010evaluation}.
Symmetry indices (SI) have also been developed, from GCF data \cite{sadeghi2000symmetry,JRNL:sant2011new}.

Balance or walking stability is another important parameter that needs to be quantified, and used to predict falls.
In~\cite{JRNL:hubble2015wearable} multiple balance and stability measures are proposed, including RMS acceleration, jerk (time series of first derivative of acceleration), sway (a measure on how much a person leans his/her body), step and stride regularity and variability.
Mobility and gait phases are also important gait parameters used to quantify gait. Mobility parameters include general movement characteristics like cadence, step length, single and double support ratio and periodicity~\cite{mizuike2009analysis, patterson2010evaluation}. Gait phases refer to the various states within one walking cycle, and there are typically eight gait phases for a healthy subject \cite{particleChase2016}.

Gait quantification can be used to extract gait features for gait pattern classification. In this paper we calculate standard gait parameters based on GCF data for mobility, balance and strength quantification. In addition, new gait phase parameters are introduced based on our previous work \cite{particleChase2016, wenlongJDSMC},
in which a wireless human motion monitoring system was
designed, and a real-time data-driven gait phase detection algorithm was developed to capture the gait phases based on the recorded GCF data. The proposed system can objectively quantify the underlying gait phases without any input from a medical professional. These two works lead to some of the gait parameters used in this paper.

\subsection{Gait pattern classification} \label{sec:relatedPatClassif}

Extensive research efforts have been reported to perform cluster analysis of post-stroke gait patterns and enable targeted treatment. In \cite{mulroy2003use} non-hierarchical cluster analysis was used to categorize four subgroups based on the temporal-spatial and kinematic parameters of walking. Similarly, hierarchical cluster analysis of post-stroke gait patterns was conducted in \cite{kinsella2008gait}, identifying three groups of patients with homogeneous levels of dysfunction. In \cite{ferrante2011biofeedback}, k-means clustering was used to group gait patterns in order to optimize participant selection in a biofeedback pedaling treatment.

Classification of post-stroke and PD gait patterns is another example of using machine learning methods in gait analysis. Classification of post-stroke gait patterns against healthy gait was performed in \cite{JRNL:chen2005gait} and \cite{mizuike2009analysis}, using kinematic and kinetic data.
Artificial neural networks (ANN) were used in \cite{kaczmarczyk2009gait} to classify post-stroke patient's gait into three categories based on the types of foot positions on the ground at first contact: \textit{forefoot, flatfoot,} and \textit{heel}. The work in \cite{chantraine2016proposition} classified hemiparetic gait in three groups with two subgroups each, that were defined from clinical knowledge. This classification method had the advantage of great usability in clinical routines without necessitating complex apparatus.
Classification of PD gait patterns against healthy gait is also studied \cite{JRNL:tahir2012parkinson,JRNL:lee2012parkinson,JRNL:wu2010statistical}. Gait features from wavelet analysis and kinematic parameters are extracted, which are passed to support vector machines (SVM) and artificial neural networks (ANN) for classification.

\eat{
	Given the remarkable diversity of gait deviations observed in post-stroke patients,
	most of the research efforts focus on studying a limited set of gait abnormalities and thus related gait parameters. For example, in \cite{kinsella2008gait} only patients with reduced knee flexion participated in the study and in \cite{kaczmarczyk2009gait} focus was only given to the subject's foot position on the ground. Extending those works to support classification of a broader set of gait disorders is very challenging. The currently used gait parameters need to support the classification of new disorders and provide statistical evidence in validating the differences between groups. Close cooperation with physical therapists and medical professionals is needed to design and select appropriate gait parameters.}
\eat{
	Also, as pointed in \cite{chantraine2016proposition}, studies with clinically driven classifications use a reduced number of inputs and define groups of patients from clinical knowledge. Despite the output of clinically meaningful groups, such studies are hightly guided by the clinical experience of only a few users. Fusion of multiple therapist's evaluation is critical to improve model accuracy.}
\eat{
	Extensive research efforts have been reported to perform classification of gait disorders, but there is no broadly accepted classification system. Such system would assist health professionals when they communicate about patients with gait disorders and it would also help in providing more targeted treatment. Research will also benefit from a good classification system, for example to ascertain that properly diagnosed patients and homogeneous groups are included in trials or gait experiments \cite{snijders2007neurological}. In this subsection we discuss smart gait pattern classification and cluster analysis methods, with specific focus on patients suffering from stroke and PD.
	
	In \cite{JRNL:tahir2012parkinson} support vector machines (SVM) and artificial neural networks (ANN) were used to classify between PD and healthy gait. In their experiments, twelve PD patients and twenty healthy subjects were asked to participate in the study and GCF were measured using force plates. Their approach reached hight classification accuracy of 98.2\%.
	In \cite{JRNL:lee2012parkinson} an open dataset of 93 PD patients and 73 healthy subjects were used for classification. Comprehensive preprocessing and wavelet analysis was used to extract features before data were fed to a neural network with weighted fuzzy membership functions. Their approach had 77\% accuracy.
	A similar dataset was used in \cite{JRNL:wu2010statistical} to classify PD vs healthy gait. The available data included kinematic parameters like stride length, swing and stance intervals, etc. Their classification accuracy is 90\% and AUC is 0.95.
	
	Classification of gait patterns in post-stroke patients was reported in \cite{mulroy2003use}. Cameras with reflective markers were used to capture kinematics of the lower extremity, isometric torques and angle at the knee, and EMG data were also recorded. Four clusters of patients were discovered, each having patients with similar gait abnormalities. Clinicians can use the critical parameters identified to categorize patients. In a similar fashion, categorization of gait patterns of post-stroke patients with equinus deformity of the foot was conducted in \cite{kinsella2008gait}. They objectively identified three subgroups of patients, where each group experienced homogeneous levels of dysfunction.
	(These two studies have concluded that gait patterns post-stroke are not homogeneous, which means that classifying stroke gait as one category can be a challenging task.)
	In \cite{JRNL:chen2005gait} a large set of gait differences were observed between hemiparetic and non-disabled control subjects at matched speeds. Kinematic and insole pressure data were collected for this task. In \cite{mizuike2009analysis} accelerometry data were used to extract gait parameters from a group of post-stroke subjects. The parameters were significantly different from a control group of healthy subjects, suggesting that accelerometry gait parameters can discriminate between the stroke patients and the control group. Artificial neural networks (ANN) were used in \cite{kaczmarczyk2009gait} to classify post-stroke patient's gait into three categories based on the type of foot position on the ground at first contact: \textit{forefoot, flatfot } and \textit{heel}. In \cite{chantraine2016proposition} they intended to put forward a new gait classification for adult patients with hemiparesis in chronic phase, and to validate its discriminatory capacity. Gait is classified in three groups with two subgroups each, that were defined from clinical knowledge, by focusing on abnormalities leading to an increase of the risk of falls.}

In this paper, we perform classification of gait patterns in three classes, healthy, Parkinson's and post-stroke.
To the best of our knowledge, there is no research work on classification of gait patterns between these three classes. We employ a comprehensive set of gait parameters
- including mobility, balance, strength and gait phases - and send them 
as input features to a classifier. An advanced classification method, MTFL, is used to distinguish between the three gait classes. Before we discuss the details of our algorithmic framework, we first present our smart shoe design for GCF data collection.

\section{Smart Shoe Design and Ground Contact Force (GCF) Data} \label{sec:smartShoes}
\begin{figure}
	\centering
	\includegraphics[width=.75\columnwidth,trim={1cm, 9cm, 0cm, 0cm},clip]{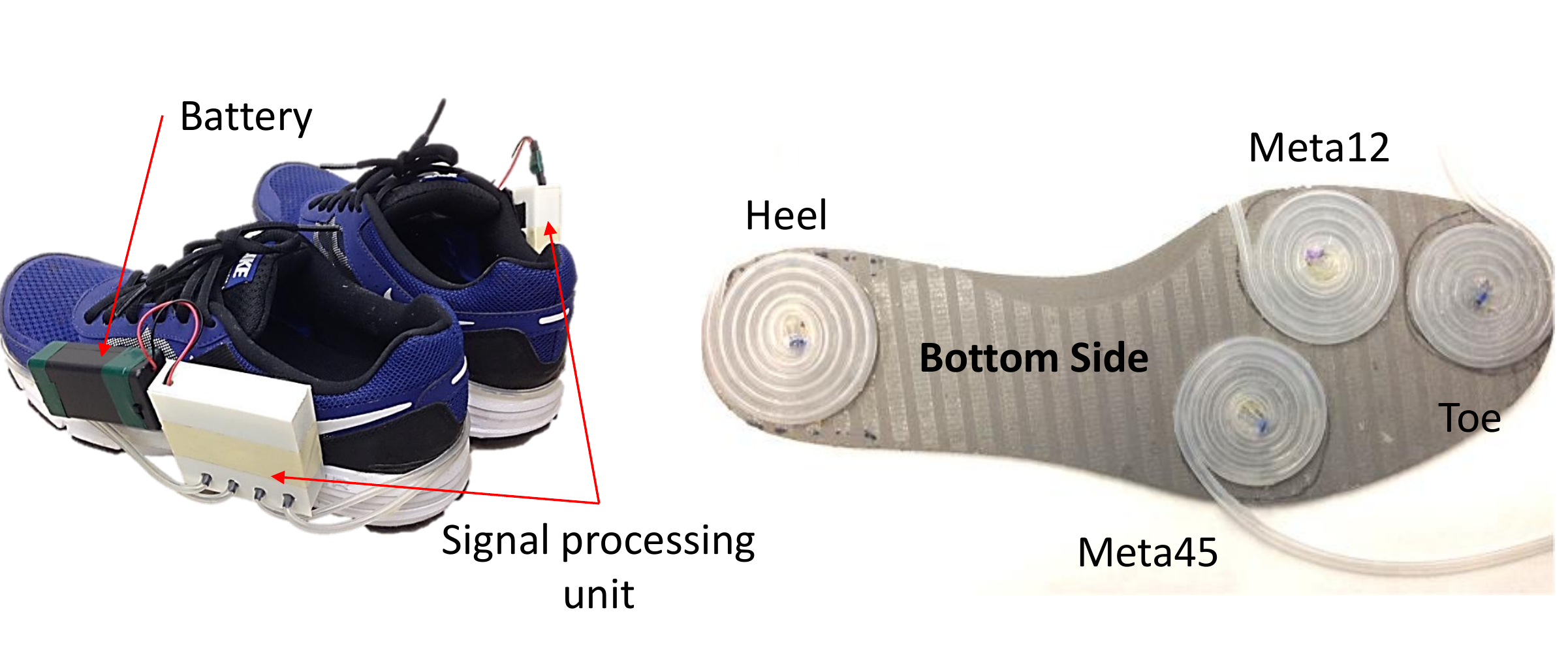}
	\caption{\small An overview of the smart shoe design. A signal processing unit includes barometric sensors, microcontroller, and Bluetooth chip}
	\label{fig:SS}
	\vspace{-.5em}
\end{figure}

In order to better analyze patients' gaits during walking, we have developed a pair of
smart shoes to measure the GCFs on both feet~\cite{wenlongJDSMC,JRNL:smartshoes}. Fig.~\ref{fig:SS} gives an overview of the shoe design. Four
barometric sensors are employed to measure the GCFs on the toe, the
first and second metatarsophalangeal (MTP) joint (Meta12), the fourth and
fifth metatarsophalangeal joint (Meta45), and the heel. Silicone
tubes are wound into air bladders to connect barometric sensors with
measurement ranging from 0 to 250 mbar. Each sensor can measure
weight up to 200 lbs with a resolution of 0.2 lbs.

The pressure sensor outputs are read by a microcontroller through
analog input channels and the sensor signals are sent out to a laptop or mobile device using a
Bluetooth module. The Bluetooth module can reliably
transmit signals in a range of 200 feet, which is enough
for normal clinical and daily use. A 9-volt alkaline battery is used
to power the smart shoes, and it can work consecutively for 90
minutes. The sampling rate of the smart shoes can go up to 100 Hz
with the Bluetooth module. In this paper the sampling rate is set at 20 Hz.
Fig.~\ref{fig:rawSS} presents the representative raw data from a healthy subject, a PD patient and a post-stroke patent, respectively. 
For the healthy subject, 
a gait cycle always starts with a strong heel strike, and then the subject moves the center of pressure to the forefoot before toe-off. Moreover, the  subject is able to maintain a good balance by allocating equal or more force to the medial boarder (Meta 12) in most of the gait cycles. However, for the PD patient, 
more force is observed on the lateral boarder (Meta 45) during the stance phase and this will significantly increase the risk of instability and falling. The stroke gait 
is even more abnormal, primarily due to the lack of heel strike as well as the poor stability shown by the large force on Meta 45. Additionally, the stroke patient walked much slower as it took 7 seconds to complete 3 steps, while the other two groups completed 5 steps in less time.

\begin{figure}
	\centering
	\includegraphics[width=\columnwidth]{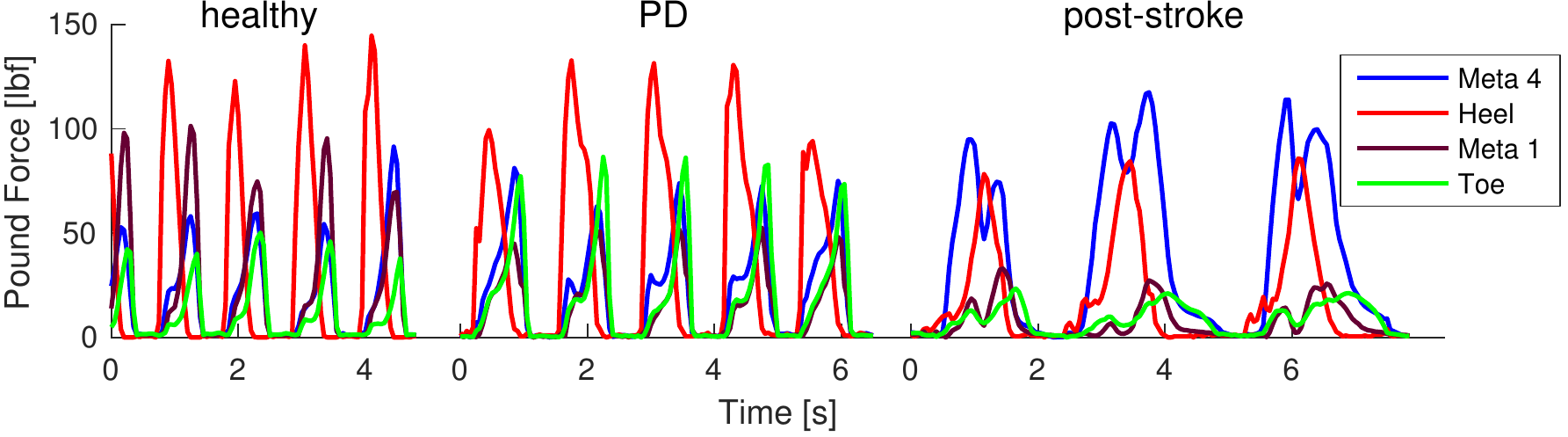}
	\caption{\small GCF data from PD and post-stroke patients and a healthy subject}
	\label{fig:rawSS}
	\vspace{-1.5em}
\end{figure}

\eat{
	\begin{figure}[htp]
		\centering
		\subfloat[\small GCF data from a healthy subject]{
			\hspace{-0.15in}
			\includegraphics[width=.45\columnwidth]{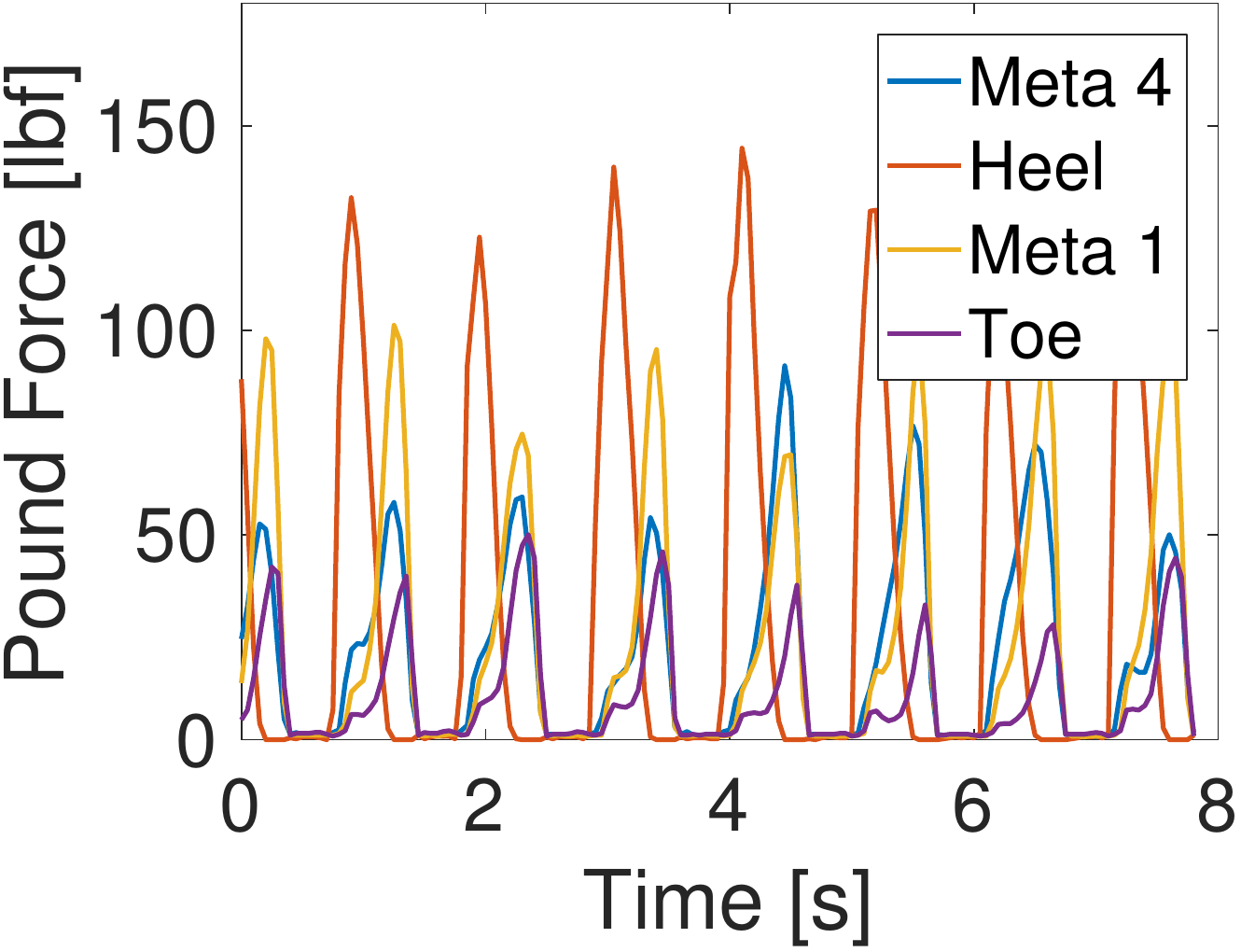}
			\label{fig:raw-healthy}}
		\hspace{.05in}
		\subfloat[\small GCF data from a PD subject]{
			\includegraphics[width=0.45\columnwidth]{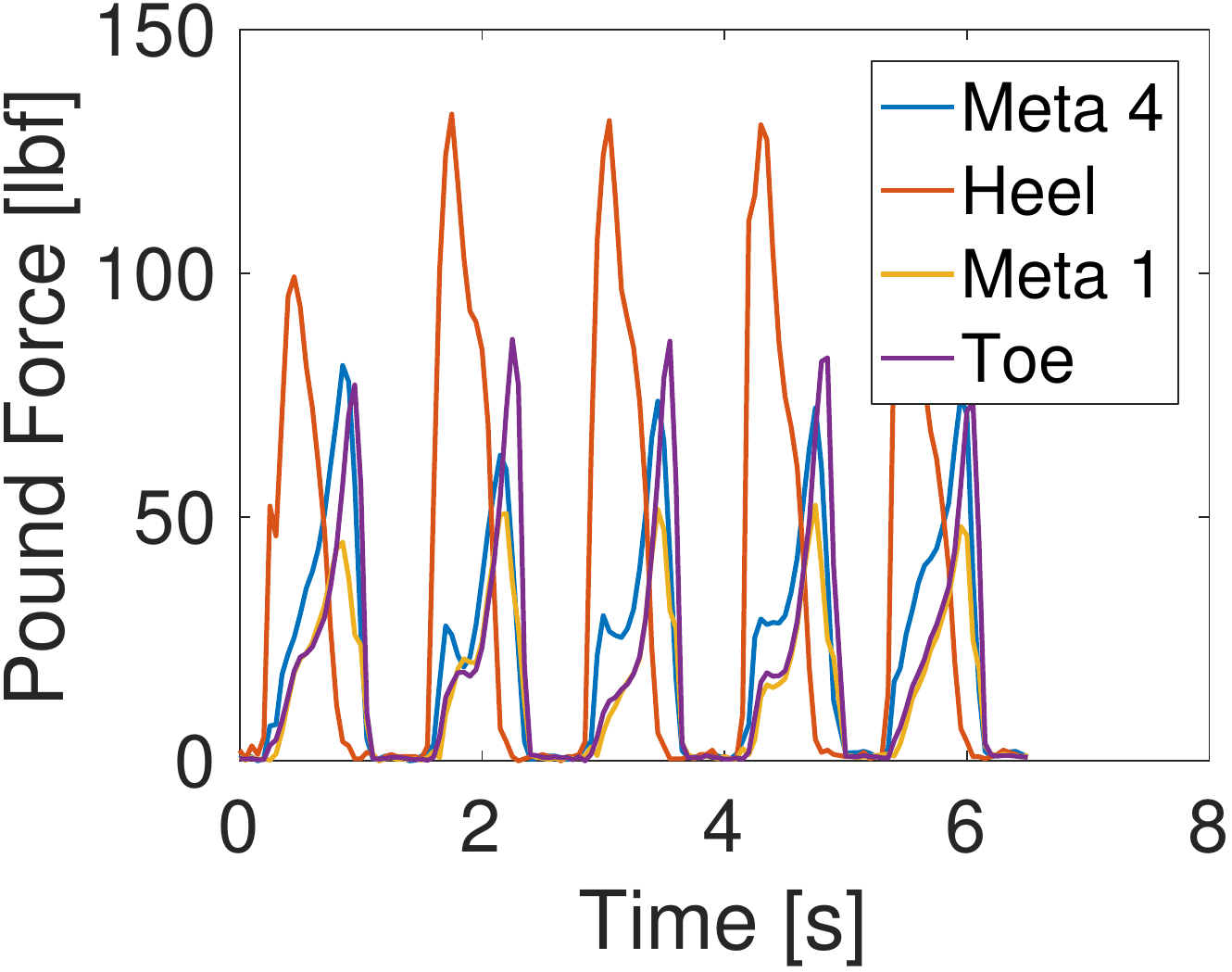}%
			\label{fig:raw-pd}}\\
		\vspace{-0.1in}
		\subfloat[\small  GCF data from a post-stroke patient]{
			\includegraphics[width=0.45\columnwidth]{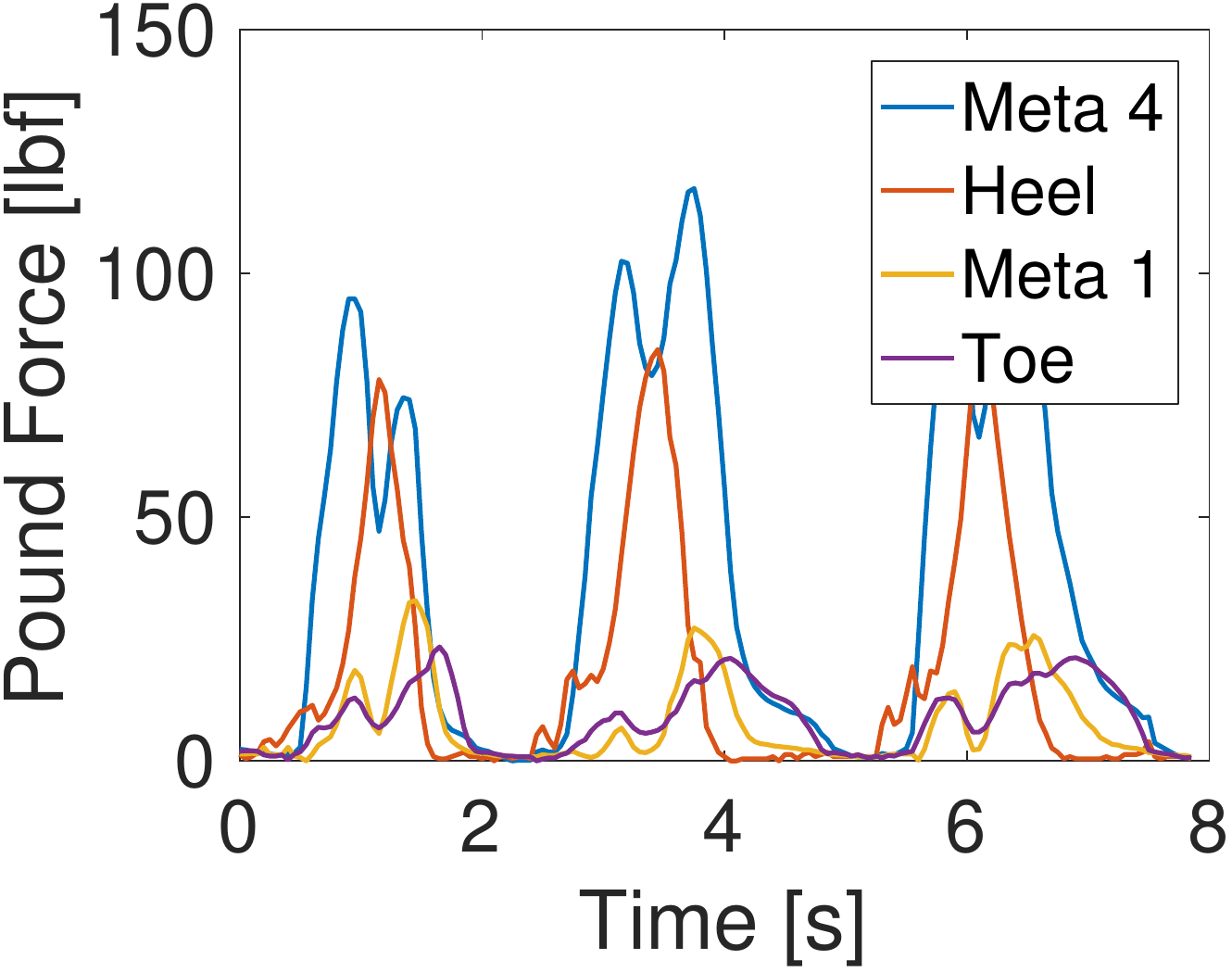}%
			\label{fig:raw-stroke}}
		\caption{\small Representative raw GCF data from the three classes of gaits, i.e. healthy subjects, PD and Post-Stroke patients}
		\vspace{-0.15in}
		\label{fig:rawSS}
	\end{figure}
}
\section{Gait Features Extraction} \label{sec:gaitParams}

To accurately describe specific human gait disorders 
is often a difficult task~\cite{snijders2007neurological}. Consequently, it is challenging to devise gait features \footnote{In the remainder of this paper we refer to gait parameters, the term used in most literature studies, as gait features to avoid confusion with the model parameters used in the multi-task learning methods in Section~\ref{sec:mtl}.} that can be used to classify gait patterns. Furthermore, the GCF data collected from the smart shoes can be noisy due to imperfect sensor dynamics and complexity of human gait. In this section we present a set of gait features that are used to detect the gait abnormalities by capturing the key gait characteristics of post-stroke and PD patients.

In Table~\ref{tab:gaitparameter}, fourteen gait features are
proposed based on the GCF data collected from the smart shoes. These featues are organized into four categories: gait phases, mobility, balance and strength. Their details will be discussed in the following subsections. Among these features, double support
ratio, single support ratio and cadence are comprehensive features, which require bilateral
information. All the other features are unilateral, as they can be calculated for each side
separately~\cite{JRNL:gaitpara}.

\begin{table}
	\centering
	\begin{tabular}{ccc}
		\hline
		Category & Gait Features & Laterality\\
		\hline
		\multirow{4}{*}{Gait Phases}& Exp. Num. of Gait Phases & Unilateral \\
		& Symmetry of Gait Phases & Unilateral \\
		& Num. of Swing Phases & Unilateral \\
		& Symmetry of Swing Phases & Unilateral \\
		\hline
		\multirow{5}{*}{Mobility} & Cadence (steps/min) & Bilateral \\
		&Double Support Ratio  & Bilateral \\
		& Single Support Ratio & Bilateral \\
		&Stance Phase Ratio  & Unilateral \\
		\hline
		\multirow{3}{*}{Balance} & \tabincell{c}{Max. Force Difference \\between Meta12 and 45}  & Unilateral \\
		&\tabincell{c}{Min. Force Difference \\between Meta12 and 45}  & Unilateral \\
		\hline
		\multirow{2}{*}{Strength}&Max. Force of Heel Strike  & Unilateral \\
		&Max. Force of Toe Off  & Unilateral \\
		\hline
	\end{tabular}
	\caption{Proposed twelve gait features in four categories}
	\label{tab:gaitparameter}
	\vspace{-1em}
\end{table}

\subsection{Gait Cycles}
\begin{figure}
	\centering
	\includegraphics[width=\linewidth]{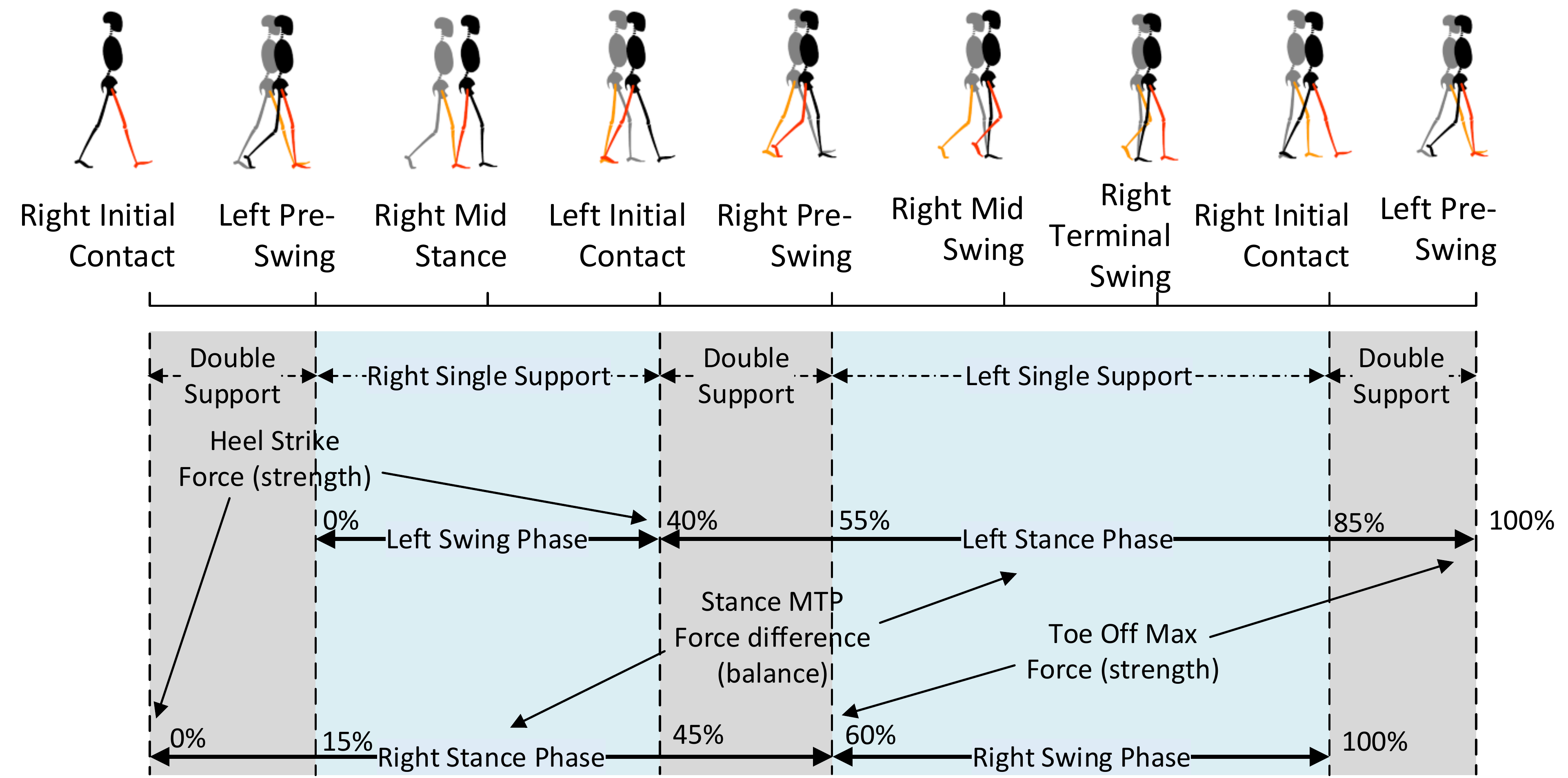}
	\caption{An overview of a gait cycle and the gait features from four categories: gait phases, mobility, balance and strength. }
	\label{fig:gaitCycle}
	\vspace{-1.5em}
\end{figure}

We first give an overview of what a gait cycle is, since all the gait features are extracted once for each gait cycle in a walking trial.  Gait cycle is the time interval between the same repetitive events of walking.  The defined cycle can start at any moment, but generally begins when one foot contacts the ground. If it starts with the right foot contacting the ground, the cycle ends when the right foot makes contact again.
Fig.~\ref{fig:gaitCycle} gives an overview of two gait cycles at the lower two horizontal solid lines. The gait cycle can be broadly divided into two phases: stance phase and swing phase \cite{BOOK:delisa1998gait}. These two phases can then be further divided into sub-phases within the gait cycle, as shown at the top part of Fig.~\ref{fig:gaitCycle}. In general, the stance phase takes around 60\% of the gait cycle \cite{BOOK:delisa1998gait} and can be divided into double support and single support. In double support, both feet are in contact with the ground, while in single support only one foot is in contact with the ground. Double or single support ratio refers to the portion of time within a gait cycle someone spends in double or single support respectively. The swing phase is described when the limb is not weight bearing and represents around 40\% of a single gait cycle \cite{BOOK:delisa1998gait}. These percentages can change with the walking speed, as with a higher speed the double support ratio in the gait cycle tends to be reduced. In Fig.~\ref{fig:gaitCycle} the lower depicted cycle starts with right foot initial contact, which leads to the stance phase, while the other starts with left pre-swing phase which leads to swing phase. Indicative percentages are shown to indicate the different phases within the cycle.

In Fig.~\ref{fig:gaitCycle}, different gait features are shown for different categories, like mobility, balance, strength. Gait pahses are shown at the top of the figure. In Sections~\ref{sec:gaitPhaseDetection} and \ref{sec:gpf} we discuss how gait phases are extracted and what gait phase related features are used in this work for gait disorder diagnosis. In Sections \ref{sec:mobility} and \ref{sec:balancestrength} we discuss other features related to mobility, balance and strength.

\subsection{Gait Phase Detection} \label{sec:gaitPhaseDetection}
Gait phases refer to various states within one walking cycle, and there are typically eight gait phases for a healthy subject (as shown at the top of Fig.~\ref{fig:gaitCycle}): initial contact, loading response, mid-stance, terminal stance (or initial contact), pre-swing, initial swing (not shown in Fig.~\ref{fig:gaitCycle}), mid-swing, and terminal swing~\cite{wenlongJDSMC,JRNL:smartshoes}.
Pathological gait can be unpredictable and complex, thus some gait phases might be missing and the time allocation of gait phases might also be different from a normal gait. This abnormal gait phase allocation provides a powerful tool for abnormal gait detection.

In this work we extract gait phase related features based on our previous work which 
applies infinite Gaussian
mixture modeling, a non-parametric Bayesian method, for gait phase detection~\cite{particleChase2016}. Our approach estimates the unknown number of gait phases that can be best described from the GCF data. Particle filters and the popular chinese restaurant process (CRP) are used for online model parameters estimation. In the rest of this subsection we describe how swing and stance phases are identified from the extracted gait phases.

\eat{
	The detection of gait phases based on the GCF measurements is essentially a classification problem. Gaussian mixture model (GMM) is a powerful tool for classification. However the total number of gait phases $K$ and the \textit{a priori} probability need to be predetermined before applying GMM, which makes it difficult to apply in actual clinical environment. Moreover, it is impossible to detect gait phases using a single ``best'' model due to the complexity of human gait. In such case, it could be useful to report the gait phase detection as a result of an average of multiple models, which means the model parameters for the GMM are also random variables. One popular way of doing that is to use Bayesian approach as follows: $p\left ( \mathbf{M}|\mathbf{y} \right )\propto p\left ( \mathbf{y}|\mathbf{M} \right )p\left(\mathbf{M} \right )$, where $p\left ( \mathbf{M}|\mathbf{y} \right )$ is the posterior probability of a model $\mathbf{M}$ given a set of measurement $\mathbf{y}$, and $p\left ( \mathbf{y}|\mathbf{M} \right )$ is the likelihood of observation $\mathbf{y}$ given the model $\mathbf{M}$.
	
	The IGMM occurs as $K\rightarrow \infty$, which means there could be infinite choices of gait phases, which is a great representation of the complexity and variability of human gait. \emph{Dirichlet} is used for the prior of the gait phases and \emph{Gaussian} for the data observations. To sample from the IGMM we use the popular method of the Chinese restaurant process (CRP)~\cite{particleChase2016}.
	
	A parallel particle filter algorithm is used for estimating the distribution of the parameters of the model, including the number of gait phases and their multivariate Gaussian distribution parameters. In a particle filter, a set of $N$ weighted ``particles", $\{C_{\{1:N\}}^{(t)},w_{\{1:N\}}^{\{t\}}\}$ (samples and associated weights at time $t$), are used to form a discrete representation of the distribution of interest (posterior distribution over class identifiers given observations). The algorithm framework comprises two main steps, {\em sampling} and {\em resampling}. The algorithm inputs are the data observations $y^{(1:T)}$ and the initial gait phases for the first $T_0$ datapoints. The algorithm then calculates the rest of the gait phases for datapoints $T_0+1$ to $T$ ({\bf $T_0$ + T?}) sequentially.
	\eat{The output of the gait phase detection algorithm can be seen in Fig.~\ref{fig:gaitPhases}.
		
		\begin{figure}[htp]
			\centering
			\includegraphics[width=0.97\columnwidth]{gaitphasespfgaits2}
			\caption{Gait phases detected using parallel particle filter}
			\label{fig:gaitPhases}
	\end{figure}}
	
	\subsection{Find Swing Phases} \label{sec:findSwing}
}

Identifying swing phases from the unlabeled gait phases is important as many other gait features are based on it.
Although it is straightforward to find healthy gait's swing phase (Fig. \ref{fig:gaitCycle}), the swing phase detection in pathological gait can be challenging for multiple reasons. First of all, the way smart shoes are worn can affect the raw GCF sensor signals. Tight shoe laces will change the raw values recorded by the barometric sensor, leading to different absolute values even for the same person in different sessions. Additionally, the stochastic nature of the sampling, which is used to estimate the distribution of gait phases~\cite{particleChase2016}, can sometimes introduce new gait phases, which are not eventually represented in the GCF data. Finally, pathological gait can be so complex that sometimes new gait phases are explored from the particle filter algorithm. Apart from that, various conditions of neural or muscular impairments, like foot-drop, can cause fore-foot dragging on the ground~\cite{BOOK:delisa1998gait}. In such cases new gait phases are likely to be discovered and they should be identified as swing phases. Having correctly identified swing phases is very important as many other features are based on them.

\eat{
	For all these reasons it is critical to identify which of the discovered gait phases are corresponding to swing phase. We need this information because many gait features to be discussed later, including double and single support ratio, need to be evaluated in only swing or stance phases. For gait phases related to foot-drop we still consider them as swing phases, as we don't want to have the other gait features affected.
}

As we described earlier, 
the swing phase ratio (portion of time spent on swing phase) of a healthy gait
is typically around 40\% of the gait cycle~\cite{BOOK:delisa1998gait}. This may change depending on the walking speed. Pathological gait can have smaller swing phase ratio, as the patient is walking slowly. Also, in the swing phase, GCF measurements will take very small positive values (or zero), as pressure from the body is not present in that limb. Using these two properties we identify the swing phases from all discovered gait phases according to the following steps:

We first calculate the average euclidean distance for all the observations in each gait phase from 0, by taking its 2-norm. We then sort the gait phases in increasing order based on their norms. We create a new swing phase, and add the observations in the sorted gait phase list one by one until the total number of observations in the new swing phase is more than 10\% of all the observations. The 10\% threshold is empirically chosen and gives the desired swing phase ratio in our dataset. The number of swing phases that were merged is kept as it is used as a gait phase feature (see Sec.~\ref{sec:gpf}). All the extracted gait phase features are described in the following subsection.

\eat{
	\begin{algorithm}
		\caption{Find Swing Phases}
		\label{alg:findSwing}
		\begin{algorithmic}[1]
			\STATE Calculate the average distance of all the observations in each gait phase from 0, by taking its 2-norm
			\STATE Sort the gait phases in increasing order based on the norms
			\STATE First gait phase is swing
			\WHILE{swing phases less than 10\% of all gait phases}
			\STATE Next gait phase is also swing
			\ENDWHILE
		\end{algorithmic}
\end{algorithm}}

\subsection{Gait Phase Features} \label{sec:gpf}
The gait phase features
are calculated from the gait phases that are extracted by our gait phase detection
algorithm (see Section \ref{sec:gaitPhaseDetection} and for more details, please refer to \cite{particleChase2016}). The expected number of gait phases can be calculated from the particles and their weights returned from the particle
filter algorithm as $\bar{K}=\sum_{i=1}^N w_i K_i $, where
$K_i$ is the number of gait phases detected from particle $i$ and $w_i$ is the particle's weight.
$\bar{K}$ is a measure of complexity of the human gait. Compared with the eight standard gait phases of a healthy subject, pathological gait is unpredictable and it may have a different number of gait phases. For example, post-stroke patients with affected neurological system may experience foot-drop. This usually increases the stance phase with circumduction to allow toe clearance~\cite{BOOK:delisa1998gait}, which can lead to toe dragging on the ground, and thus causing the gait phase detection algorithm detecting multiple swing phases. The number of swing phases is another gait parameter and has been discussed in the previous Section~\ref{sec:gaitPhaseDetection}.

The symmetry of gait phases (swing phases) is used as a
measure to quantify how even the proportion of time spent is in each gait phase in a gait cycle (swing phases). We chose to include this new type of symmetry measure as it can be easily applied on the gait phases that were extracted from our Dirichlet process mixture model~\cite{particleChase2016}, given the fact that number of gait phases is not known \textit{a-priori} for each subject. This single gait parameter can estimate the symmetry for any number of gait phases detected.
It is based on the cosine similarity, as described in the following formula:
\begin{equation}
cos(\theta) = \frac{\vect g \cdot \vect u^T}{||\vect g|| \cdot ||\vect u||}
\end{equation}
where $\theta$ is the angle between $\vect g$ and $\vect u$, with $\vect g, \vect u \in \mathbb{N}^K$ and $K$ is the number of gait phases (swing phases) found. $\vect g$ is a vector of size $K$, where each element in $\vect g$ counts the number of observations belonging to each gait phase (swing phase) within a gait cycle and $\vect u$ is a vector of size $K$ with all its elements equal to 1, assuming observations in $\vect u$ are evenly distributed.
If the number of observations belonging to each gait phase is not evenly distributed and thus there are gait phases with
very few observations, the angle between vector $\vect g$ and $\vect u$ will be higher
resulting in lower symmetry. On the other hand, if the number of observations belonging to one gait phase is always more than normal it would also result in lower  symmetry.

\subsection{Mobility Features} \label{sec:mobility}
We select four features in the mobility category, cadence, double and single support ratios and stance phase ratio. Cadence is measured in steps per minute and it is calculated by taking the total number of stance phases in one trial divided by the length of the trial in minutes.
The double support ratio refers to
the proportion of time in a gait cycle that both feet are in the
stance phase to support the subject, whereas the single support
ratio refers to the proportion of time in a gait cycle that only one
foot touches the ground while the other is in the swing phase. Stance
phase ratio refers to the proportion of time in a gait cycle that
one foot is in the stance phase. All these features are summarized in Fig. \ref{fig:gaitCycle}.

\subsection{Balance and Strength Features} \label{sec:balancestrength}
We select two features in the balance and strength categories each. In the balance category, the maximum and minimum force differences between the medial (Meta12, Fig.~\ref{fig:SS}) and
lateral (Meta45, Fig.~\ref{fig:SS}) sides of the forefoot in a gait cycle can be calculated as
\begin{eqnarray}
\max_{i\subseteq \mathbb{I}} F_{M12}(i)-F_{M45}(i),\\
\min_{i\subseteq \mathbb{I}} F_{M12}(i)-F_{M45}(i).
\end{eqnarray}
These features can evaluate the capability of maintaining balance.
The $\mathbb{I}$ refers to the set of indices $i$ that belong to one
gait cycle. Strength is quantified using the maximum force on the
heel during heel strike and on the toe during toe off. All balance
and strength features are normalized by the body weight to make
them comparable among different subjects.

\section{Multi-Task Feature Learning for Gait Disorder Diagnosis} \label{sec:mtl}

Based on the extracted gait features, we diagnose gait disorders by constructing classifiers as functions of these features. In this work, we use an advanced multi-task feature learning (MTFL) classification method~\cite{JMLR:v17:15-234} to build three classifiers to discriminate gait observations of PD and stroke patients, respectively, from those of healthy adults as well as in between the gaits of PD and stroke patients. The selected learning strategies can be more feasible to identify similarities and differences of gait patterns than classic multi-class classification algorithms given multi-class classification methods focus on modeling only the exclusive (or discriminative) features of the different gait classes. Moreover, the methodology helps in important gait feature selection which may help in better understanding the key characteristics that distinguish abnormal gaits and help design more targeted treatment methods.

MTL is a methodology that can improve the generalization of multiple related classification tasks by exploiting the task relationships, especially when the training set for some or all the tasks is limited. Related tasks are learned in a joint manner, so that knowledge learned from one task may benefit learning for other tasks. For example, in gait disorder diagnosis, the task of deciding if an observation, represented by a vector of gait features, is recorded from a PD patient or healthy subject, may help diagnose if another observation is recorded from a post-stroke patient or a healthy subject. MTL has been shown to be theoretically and practically more effective than learning tasks individually~\cite{JMLR:v17:15-234}.
A widely-used basic assumption is that the related tasks may share a common representation in the feature space, which is investigated by multi-task feature learning (MTFL).

We revisit two of our recently developed MTFL methods that both rely on a multiplicative decomposition of the model parameters used for each task, and hence are referred to as Multiplicative MTFL (MMTFL). Both methods are related to the widely used block-wise joint regularization MTFL method \cite{obozinski2006multi}, but bring out a significant advantage over it, in terms of selecting relevant features for classification. The new methods can simultaneously select features that are useful across multiple tasks and features that might be only discriminative for a specific classification task.

Given $T$ classification tasks in total, let $(\matrx X_t \in \mathbb{R}^{\ell_t \times d}, \matrx y_t \in \mathbb{R}^{\ell_t})$ be the sample set for the $t$-th task, where $\matrx X_t$ is a matrix containing rows of examples and columns of gait features, $\vect y_t$ is a column vector containing the corresponding labels for each example, $\ell_t$ is the sample size of task $t$, and $d$ is the number of features. We focus on creating linear classifiers $\vect y_t = sign(\matrx X_t \vect \alpha_t)$, where $\matrx \alpha_t$ is the vector of model parameters to be determined. We then define a model parameter matrix $\matrx A$ where each column contains a task's parameter vector $\vect \alpha_t$, and thus each row of this matrix corresponds to a gait feature, i.e., the weights for a gait feature used for each of the $T$ tasks, which we denote as $\vect \alpha^j$, and $j = 1,\cdots,d$. We choose a loss function $L(\matrx \alpha_t, \matrx X_t, \matrx y_t)$ which typically measures the discrepancy between the prediction $\matrx X_t \vect \alpha_t$ and the observation $\vect y_t$ for task $t$. In a classification task, the loss function is commonly a logistic regression loss.

The widely used block-wise joint regularization MTFL method solves the following optimization problem for the best $\vect \alpha$:
\begin{equation} \label{eq:generalReg}
\min_{\matrx \alpha_t}\sum\limits_{t=1}^TL(\matrx \alpha_t, \matrx X_t, \matrx y_t) + \lambda \Omega(\matrx A), \quad t=1,\cdots,T,
\end{equation}
where $\Omega(\matrx A)$ is a block-wise regularizer, often called the $\ell_{1,p}$ matrix norm, that computes $\sum_{j=1}^d ||\vect \alpha^j||_p$. Common choices for $p$ are $1$, $2$ or $\infty$. Minimizing this $\ell_{1,p}$ regularizer can shrink an entire row of $\matrx A$ to zero, thus eliminating or selecting features for all tasks. The hyperparameter $\lambda$ is used to play the trade-off between the loss function and the regularizer. However, a major limitation of the joint regularization MTFL method is that it either selects a feature for all tasks, or eliminates it from all tasks, which can be unnecessarily restrictive. In practice, several tasks may share features but some features may only be useful for a specific task. Hence, we introduce the following multiplicative MTFL that addresses this issue.

A family of MMTFL methods can be derived by factorizing $\matrx \alpha_t = \vect c \odot \matrx \beta_t$, where $\odot$ computes a vector whose $j$-th component equals the product of $\vect c_j$ and $\vect \beta_t^j$, and in other words, $a_t^j=c_j\beta_t^j$.  The vector $\vect c$ is applied across tasks, indicating whether certain features are useful to any of the tasks, and $\vect \beta_t$ is only relevant to task $t$. We relax the indicator vector $\vect c$ (i.e., a binary vector) into a non-negative $\vect c$ so the optimization problem can be tractable. If $c_j=0$, then the $j$-th feature will not be used by any of the models. If $c_j > 0$, then a specific $\beta_t^j=0$ can still rule out the $j$-th feature from the $t$-th task. We minimize a regularized loss function with separate regularizers for $\vect c$ and $\vect \beta_t$ as follows for the best models:
\begin{equation}\begin{array}{@{}ll}
\min\limits_{\vect \beta_t,\vect c \ge 0} &
\sum\limits_{t=1}^T L(\matrx c,\vect \beta_t,\matrx X_t,
\matrx y_t) + \gamma_1\sum\limits_{t=1}^T||\vect \beta_t||_p^p+\gamma_2||\vect c||_k^k,\\
\end{array} \label{eq:algform_MMTFL}
\end{equation}
where $||\vect \beta_t||_p^p=\sum_{j=1}^{d}|\beta_t^j|^p$ and $||\vect c||_k^k=\sum_{j=1}^{d}(c_j)^k$, which are the $\ell_p$-norm of $\vect \beta_t$ to the power of $p$ and the $\ell_k$-norm of $\vect c$ to the power of $k$ if $p$ and $k$ are positive integers. The tuning parameters $\gamma_1$, and $\gamma_2$ are used to balance the empirical loss and regularizers. According to the different choices of $p$ and $k$, we can have different levels of sparsity for $\vect c$ and $\vect \beta_t$.

The method MMTFL(2,1) refers to the case when $p=2$ and $k=1$ in Eq.(\ref{eq:algform_MMTFL}) and solves a problem as follows:
\begin{equation}\begin{array}{@{}ll}
\min\limits_{\vect \beta_t,\vect c \ge 0} &
\sum\limits_{t=1}^T L(\matrx c,\vect \beta_t,\matrx X_t,
\matrx y_t) + \gamma_1\sum\limits_{t=1}^T||\vect \beta_t||_2^2+\gamma_2||\vect c||_1,\\
\end{array} \label{eq:algform_MMTFL21}
\end{equation}
It is widely known that $\ell_2$-norm is not sparsity-inducing, meaning that minimizing it leads to a vector of many small, non-zero entries. On the other hand, the sparsity-inducing $\ell_1$-norm creates a vector with many entries equal to zero. In Eq.(\ref{eq:algform_MMTFL21}), $\vect c$ is regularized by a sparsity-inducing norm, hence tending to eliminate many features from across all of the tasks. This formulation is more suitable for capturing the feature sharing pattern such that there exists a large subset of irrelevant features across tasks, requiring a sparse $\vect c$, but different tasks share a significant amount of features from the selected feature pool as indicated by $\vect c$, thus requiring a non-sparse $\vect \beta_t$.

The method MMTFL(1,2) is on the opposite direction when $p=1$ and $k=2$ in Eq.(\ref{eq:algform_MMTFL}), and solves the following problem:
\begin{equation}\begin{array}{@{}ll}
\min\limits_{\vect \beta_t,\vect c \ge 0} &
\sum\limits_{t=1}^T L(\matrx c,\vect \beta_t,\matrx X_t,
\matrx y_t) + \gamma_1\sum\limits_{t=1}^T||\vect \beta_t||_1+\gamma_2||\vect c||_2^2.\\
\end{array} \label{eq:algform_MMTFL12}
\end{equation}
Eq.(\ref{eq:algform_MMTFL12}) is suitable to capture a feature sharing pattern where none or only a small portion of the features can be removed because each may be useful for some tasks, thus requiring a non-sparse $\vect c$. However, different tasks share a small amount of these features, thus requiring a sparse $\vect \beta_t$. In this case, $\ell_1$-norm is applied to $\vect \beta_t$ and $\ell_2$-norm is applied to $\vect c$.

Since it is difficult to prove any relationship between gait features and actual gait problems, we hypothesize that these methods can help us identify the important gait features to recognize abnormal gaits due to the neurological diseases from otherwise healthy gaits, and may further locate features to discriminate between stroke-induced gaits and PD-induced gaits. To validate this hypothesis, in our performance evaluation, we compare the two methods against early MMTFL methods that are most comparable to the proposed methods and two baseline methods - single task learning (STL) methods that either use $\ell_2$-norm or $\ell_1$-norm to regularize individual $\vect \alpha_t$, which we referred to as STL-ridge and STL-lasso respectively.

\section{Performance Evaluation}
\label{sec:expEval}

We designed two sets of experiments to evaluate the effectiveness of the proposed methods. In the first set of experiments, we examined the area under the curve (AUC) classification performance metric of the models that are created by the different MTFL methods.  In the second set of experiments we studied the importance of each proposed gait feature and their relevance to each classification task.
In the following, we first describe our human participant study design and then present the experiment details.

\subsection{Human Subject Test Design} \label{sec:clinicalStudy}
In order to evaluate the performance
of the proposed algorithms, we collected GCF data using the developed smart shoes from healthy
subjects without known walking problems and PD and post-stroke patients. Experiments with healthy subjects
were conducted in the Mechanical Systems Control Laboratory at
the University of California, Berkeley. The
clinical study with patients was conducted in the William J. Rutter
Center at the University of California, San Francisco (UCSF). The Committee
on Human Research (CHR) at UCSF reviewed and approved this
study. The original purpose of this human subject study was to examine whether patients
could use visual feedback to direct their rehabilitation training and
how was the training performance compared to traditional rehabilitation
training directed by a physical therapist only. We use these datasets to evaluate the algorithm developed in this paper. Detailed experimental
design and statistical analysis of the clinical outcomes are available
in~\cite{byl2015clinical,wenlongJDSMC}.

To collect data for this work, the subjects were asked to walk multiple trials on a flat ground for at least 50 consecutive steps in their normal walking speeds. The data collected from five PD patients, three post-stroke patients, and three healthy subjects are used to test our methodology. The average ages for each of the groups are 69.2, 53 and 23 years old respectively. Representative raw data from each of the three groups are shown in Fig.~\ref{fig:rawSS}. Gait features are extracted for each gait cycle and average results are taken for each trial. This generates a dataset of 180 observations with 21 features each.

\subsection{Classification of Gait Disorders} \label{sec:classifEval}
To classify among stroke, PD and healthy gaits we designed and evaluated 3 classification tasks: healthy v.s. stroke gait, healthy v.s. Parkinson's gait and stroke v.s. Parkinson's gait. We compared our two new formulations MMTFL\{2,1\} and MMTFL\{1,2\} with two other standard MMTFL methods. They are all summarized as follows:
\begin{itemize}
	\item MMTFL\{2,1\}: formulation (\ref{eq:algform_MMTFL21})
	\item MMTFL\{1,2\}: formulation (\ref{eq:algform_MMTFL12})
	\item MMTFL\{1,1\}: formulation (\ref{eq:algform_MMTFL}) with $p=k=1$
	\item MMTFL\{2,2\}: formulation (\ref{eq:algform_MMTFL}) with $p=k=2$
\end{itemize}

In addition, two single task learning (STL) approaches were implemented as baselines and compared with the MTFL algorithms. They can be formulated as folows:
\begin{equation} \label{eq:stlReg}
\min_{\matrx \alpha_t}\sum_i ||y_t^i -  X_t^i \vect \alpha_t || + \lambda \Omega(\matrx a_t), \quad t=1,\cdots,T,
\end{equation}
With $X_t^i$ and $y_t^i$ the i-th example and example label for task $t$ respectively, $\vect \alpha_t$ the parameter vector for task $t$, $\lambda$ the hyperparameter used to play the trade-off between the least squares loss and the regularizer and $\Omega$ the selected regularizer.
They are summarized as follows:
\begin{itemize}
	\item STL-lasso: with $||a_t||_1$ as the regularizer
	\item STL-ridge: with $||a_t||_2^2$ as the regularizer
\end{itemize}

Before we ran the experiments we used a tuning process to find appropriate values for the hyperparameters, $\gamma_1$ and $\gamma_2$. Grid search with three-fold cross validation (CV) was performed to select proper hyperparameter values in the range from $10^{-3}$ to $10^3$. In all the experiments, hyperparameters were fixed to the values that yielded the best performance in the CV.

In the first set of experiments,
we partitioned the 180 observations into a training dataset and a testing dataset according to a given partition ratio, which was set to be 16\%, 20\%, 25\%, 33\% or 50\%, respectively in each experiment.
For each partition ratio, 10-fold CV was performed and average results were reported.
The classification performance was measured using AUC, which measures the total area under the receiver operating characteristic (ROC) curves. These results are summarized in the left half of table \ref{tab:mtlResults}. We can observe from the results that MTFL methods always outperform STL methods.
Specifically, with the smallest training set of 16\%, the MMTFL\{2,1\} method has the best improvement over the STL methods.
When the training partition ratio was increased, the AUC performance of all the methods improved consistently.
When it reached 50\%, STL or MTFL methods achieved their highest AUC scores, respectively.
The advantage of MTFL methods with smaller training set ratios is explained because they can learn the tasks jointly and not exclusively, which is typically done in STL methods.
On the other hand, along with the increase of training dataset percentage, more training examples are provided to the classifiers, making the classification easier and thus STL methods performed closer to MTFL when the partition rate increases.

\begin{table*}
	\centering
	\begin{tabular}{c|ccccc}
		& \multicolumn{5}{c}{\bf Random Partition}  \\
		{\bf Method}   &   \textbf{16}\%  &   \textbf{20}\%  &   \textbf{25}\%  &   \textbf{33}\%  &   \textbf{50}\% \\
		
		\hline
		\textbf{MMTFL\{2,2\}} 	 &  0.93$\pm$0.03 	 &  0.97$\pm$0.02 	 &  0.97$\pm$0.01 	 &  0.98$\pm$0.01 	 &  0.99$\pm$0.01  \\
		
		\textbf{MMTFL\{1,1\}} 	 &  0.94$\pm$0.04 	 &  0.96$\pm$0.02 	 &  0.98$\pm$0.01 	 &  0.98$\pm$0.01 	 &  0.99$\pm$0.01 \\
		
		\textbf{MMTFL\{2,1\}} 	 &  0.95$\pm$0.03 	 &  0.97$\pm$0.02 	 &  0.98$\pm$0.01 	 &  0.98$\pm$0.01 	 &  0.99$\pm$0.01 \\
		
		\textbf{MMTFL\{1,2\}} 	 &  0.93$\pm$0.04 	 &  0.96$\pm$0.03 	 &  0.98$\pm$0.01 	 &  0.98$\pm$0.01 	 &  0.99$\pm$0.01 \\
		
		\textbf{STL-ridge} 	 &  0.90$\pm$0.03 	 &  0.94$\pm$0.03 	 &  0.95$\pm$0.02 	 &  0.97$\pm$0.01 	 &  0.98$\pm$0.01 \\
		
		\textbf{STL-lasso} 	 &  0.92$\pm$0.03 	 &  0.96$\pm$0.02 	 &  0.97$\pm$0.02 	 &  0.98$\pm$0.01 	 &  0.99$\pm$0.00 \\

	\end{tabular}
	\caption{AUC performance of different methodologies}
	\label{tab:mtlResults}
	\label{tab:mtlResultsLeaveOneOut}
	\vspace{-1em}
\end{table*}

Following that, we tested how well the classification generalizes when a new subject's gait was tested against a model built by gaits of other patients and healthy subjects. Specifically, the same classification tasks were performed with the same classification methods, but the testing data were from a single subject and all the data from the rest of subjects were used to train the corresponding model. We repeated this for each individual patient and healthy subject and the performance results are summarized in the right half of Table \ref{tab:mtlResultsLeaveOneOut}, where average AUC is reported across all tasks and per task separately. PD, ST and H refer to the gait from PD patients, post-stroke patients and healthy subjects, respectively.

\begin{table}
	\centering
	\begin{tabular}{c|c|c|c|c}
		\multirow{2}{*}{\bf Method} & All tasks & PD vs H & ST vs H & ST vs PD \\
		& AUC & AUC & AUC & AUC \\
		\hline
		
		\textbf{ MMTFL\{2,2\} }   &  0.949    &  0.880    &  0.994    &  0.967   \\
		\textbf{ MMTFL\{1,1\} }   &  0.979    &  0.982    &  0.993    &  0.960   \\
		\textbf{ MMTFL\{2,1\} }   &  0.978    &  0.960    &  0.994    &  0.979   \\
		\textbf{ MMTFL\{1,2\} }   &  0.975    &  0.983    &  0.983    &  0.967   \\
		\textbf{ STL-ridge }   &  0.916    &  0.831    &  0.971    &  0.940   \\
		\textbf{ STL-lasso }   &  0.944    &  0.893    &  0.977    &  0.961   \\
		
	\end{tabular}
	\caption{Per task average AUC scores when a new subject is tested in a model trained by the rest subjects}
	\label{tab:mtlResultsLeaveOneOut}
\end{table}
\eat{
	\begin{table*}
		\centering
		\begin{tabular}{clcccccc}
			& \textbf{Metric} & \textbf{MMTFL\{2,2\}} & \textbf{MMTFL\{1,1\}} & \textbf{MMTFL\{2,1\}} & \textbf{MMTFL\{1,2\}} & \textbf{STL-ridge} & \textbf{STL-lasso}  \\
			\hline
			\multirow{3}{*}{All tasks} & AUC & 0.949 & 0.979 & 0.978 & 0.975 & 0.916 & 0.944 \\
			& F1 score & 0.854 & 0.915 & 0.905 & 0.879 & 0.800 & 0.827 \\
			& ErrorRate & 0.110 & 0.067 & 0.073 & 0.096 & 0.154 & 0.132 \\
			\hline
			\multirow{3}{*}{PD vs Healthy} & AUC & 0.880 & 0.982 & 0.960 & 0.983 & 0.831 & 0.893 \\
			& F1 score & 0.850 & 0.963 & 0.917 & 0.940 & 0.800 & 0.831 \\
			& ErrorRate & 0.156 & 0.041 & 0.088 & 0.068 & 0.224 & 0.177 \\
			\hline
			\multirow{3}{*}{Stroke vs Healthy} & AUC & 0.994 & 0.993 & 0.994 & 0.983 & 0.971 & 0.977 \\
			& F1 score & 0.893 & 0.912 & 0.949 & 0.784 & 0.800 & 0.842 \\
			& ErrorRate & 0.063 & 0.053 & 0.032 & 0.116 & 0.116 & 0.095 \\
			\hline
			\multirow{3}{*}{Stroke vs PD} & AUC & 0.967 & 0.960 & 0.979 & 0.967 & 0.940 & 0.961 \\
			& F1 score & 0.828 & 0.794 & 0.828 & 0.794 & 0.800 & 0.800 \\
			& ErrorRate & 0.088 & 0.114 & 0.088 & 0.114 & 0.096 & 0.105 \\
			
		\end{tabular}
		\caption{Average task classification performance using leave one patient out}
		\label{tab:mtlResultsLeaveOneOut}
\end{table*}}

As can be observed from Table \ref{tab:mtlResultsLeaveOneOut}, MTFL methods performed better than STL methods consistently.
We also observe that there were some easier tasks (e.g., stroke vs healthy), where STL AUC scores were almost as good as MTFL ones, and some more challenging tasks (e.g., PD vs healthy), where STL AUC scores were worse compared to any other task.

To further study how the two new MTFL formulations perform on each task we report the confusion matrices of all the three tasks for MMTFL\{1,2\} and MMTFL\{2,1\} in Table \ref{tab:mtlL12Confusion} and \ref{tab:mtlL21Confusion} respectively. Each row in the matrix corresponds to which gait class was tested, while a column corresponds to which gait class the algorithm predicted. Between these two new formulations, MMTFL\{1,2\} performed better with PD, as out of the 83 tested gaits, MMTFL\{1,2\}  predicted 5 of them to be healthy gaits, i.e. false negatives, compared to 11 healthy gaits that were predicted by MMTFL\{2,1\}. MMTFL\{2,1\} performed better with post-stroke, as
out of the 31 tested stroke gaits MMTFL\{2,1\}  predicted 3 of them to be healthy gaits, compared to 11 healthy gaits that were predicted by MMTFL\{1,2\}. Overall, MMTFL\{2,1\} performed better, as it also achieved beter false positive rates. Specifically MMTFL\{2,1\} predicted only 2 PD gaits out of 64 healthy gaits and 3 stroke gaits out of the 83 PD gaits, compared to 5 and 7 predicted by MMTFL\{1,2\} in the same tasks respectively.

\begin{table*}
	\centering
	\begin{tabular}{c|cc}
		& PD & Healthy \\
		\hline
		PD & 78 & 5 \\
		Healthy & 5 & 59 \\
	\end{tabular}
	\hspace{1cm}
	\begin{tabular}{c|cc}
		& Stroke & Healthy \\
		\hline
		Stroke & 20 & 11 \\
		Healthy & 0 & 64 \\
	\end{tabular}
	\hspace{1cm}
	\begin{tabular}{c|cc}
		& Stroke & PD \\
		\hline
		Stroke & 25 & 6 \\
		PD & 7 & 76 \\
	\end{tabular}
	\caption{Confusion Matrices of \textbf{MMTFL\{1,2\}} for the 3 tasks, true labels in rows, predicted in columns}
	\label{tab:mtlL12Confusion}
	\vspace{-1em}
\end{table*}

\begin{table*}
	\centering
	\begin{tabular}{c|cc}
		& PD & Healthy \\
		\hline
		PD & 72 & 11 \\
		Healthy & 2 & 62 \\
	\end{tabular}
	\hspace{1cm}
	\begin{tabular}{c|cc}
		& Stroke & Healthy \\
		\hline
		Stroke & 28 & 3 \\
		Healthy & 0 & 64 \\
	\end{tabular}
	\hspace{1cm}
	\begin{tabular}{c|cc}
		& Stroke & PD \\
		\hline
		Stroke & 24 & 7 \\
		PD & 3 & 80 \\
	\end{tabular}
	\caption{Confusion Matrices of \textbf{MMTFL\{2,1\}} for the 3 tasks, true labels in rows, predicted in columns}
	\label{tab:mtlL21Confusion}
	\vspace{-1em}
\end{table*}

The last set of experiments aimed to report the prediction results per patient, in order to give complete information of the performance of each subject's gait. Table \ref{tab:mtlL21PatientConfusion} summarizes the per patient confusion matrices generated from MMTFL\{2,1\} for the three classification tasks.
The first column indicates each subject's disease or healthy condition and their identification numbers (ID) are given in the second column. The last two columns give number of times a trial was predicted to be PD, stroke or healthy subjects. The summation of these two numbers  in each row corresponds to the total number of trials that were recorded for each subject.
From the table we observe that stroke patient 4 was almost always predicted either healthy subject or PD patient, which means that her gait patterns were much different from the other post-stroke patients. This patient was a 33 year old female with minor stroke, which explains the similarity of her gait to a healthy, when compared to other older stroke patients. This wrong prediction may also be related to the limited number of stroke paitents that participated in this study.

\begin{table*}
	\centering
	\begin{tabular}{cc|cc}
		\multicolumn{2}{c}{Subject}  & \multicolumn{2}{c}{Predicted} \\
		Disease & ID & PD & Healthy \\
		\hline
		PD & 1 & 16 & 0 \\
		PD & 2 & 11 & 6 \\
		PD & 3 & 13 & 5 \\
		PD & 5 & 19 & 0 \\
		PD & 6 & 13 & 0 \\
		Healthy & 7 & 1 & 22 \\
		Healthy & 8 & 1 & 21 \\
		Healthy & 9 & 0 & 19 \\
	\end{tabular}
	\hspace{1cm}
	\begin{tabular}{cc|cc}
		\multicolumn{2}{c}{Subject}  & \multicolumn{2}{c}{Predicted} \\
		Disease & ID & Stroke & Healthy \\
		\hline
		Stroke & 4 & 4 & 3 \\
		Stroke & 10 & 8 & 0 \\
		Stroke & 11 & 16 & 0 \\
		Healthy & 7 & 0 & 23 \\
		Healthy & 8 & 0 & 22 \\
		Healthy & 9 & 0 & 19 \\
	\end{tabular}
	\hspace{1cm}
	\begin{tabular}{cc|cc}
		\multicolumn{2}{c}{Subject}  & \multicolumn{2}{c}{Predicted} \\
		Disease & ID & Stroke & PD \\
		\hline
		PD & 1 & 0 & 16 \\
		PD & 2 & 0 & 17 \\
		PD & 3 & 1 & 17 \\
		PD & 5 & 2 & 17 \\
		PD & 6 & 0 & 13 \\
		Stroke & 4 & 1 & 6 \\
		Stroke & 10 & 7 & 1 \\
		Stroke & 11 & 16 & 0 \\
	\end{tabular}
	\caption{Confusion Matrices of \textbf{MMTFL\{2,1\}} for the 3 tasks per patient}
	\label{tab:mtlL21PatientConfusion}
	\vspace{-1.5em}
\end{table*}

Given that MMTFL\{2,1\} performs best in general, the tested data seem to follow the assumption under which MMTFL\{2,1\} was designed. Specifically, across all three tasks there exists a large subset of irrelevant sensing features, requiring a sparce $\vect c$, but different tasks share a significant amount of features from the selected fetature pool as indicated by $\vect c$.  In other words, there are some specific sensing features that help identify the neurological disorders. In the next subsection we are going to present the selected features for each method used in this paper.

\subsection{Identification of Important Gait Features}
\begin{figure}
	\centering
	\includegraphics[width=\columnwidth]{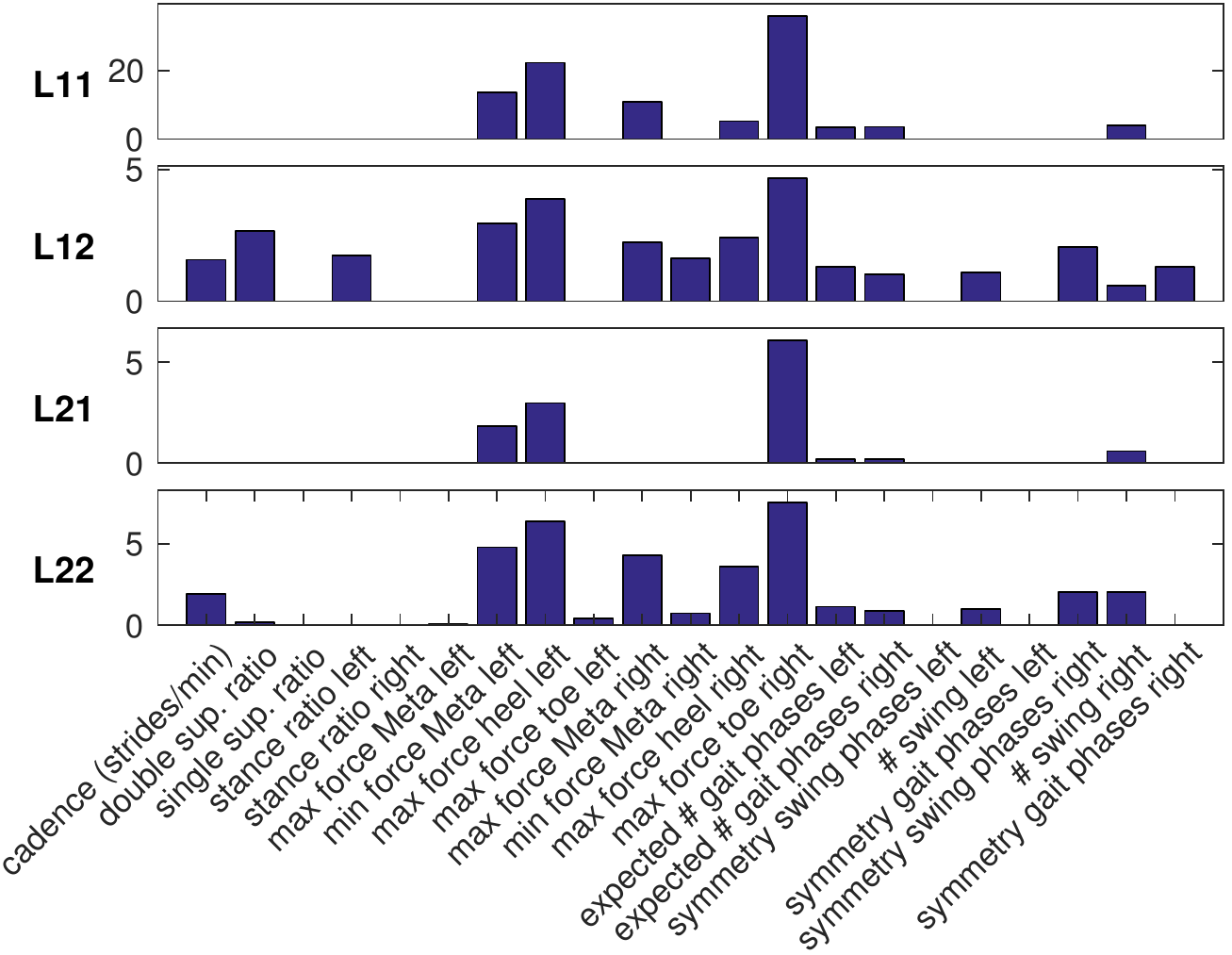}
	\caption{Feature selection vector $\vect c$ from all MMTFL methods}
	\label{fig:featuresSelected}
\end{figure}

\begin{figure}
	\centering
	\includegraphics[width=\columnwidth]{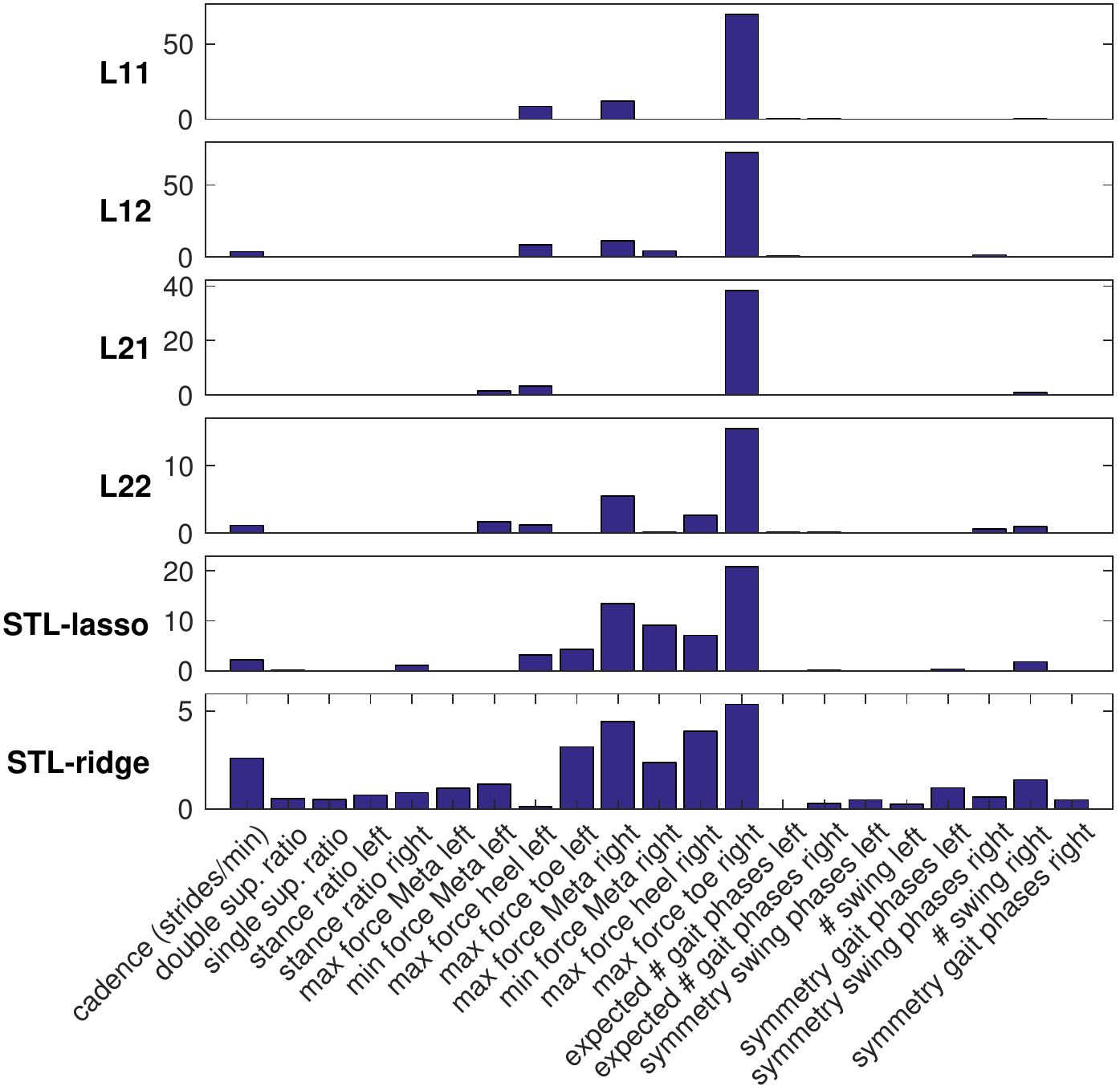}
	\caption{Absolute value of task parameter vector $\vect \alpha_t$ in the PD vs. healthy gait classification task.}
	\label{fig:pdvsh}
\end{figure}

\begin{figure}
\centering
\includegraphics[width=\columnwidth]{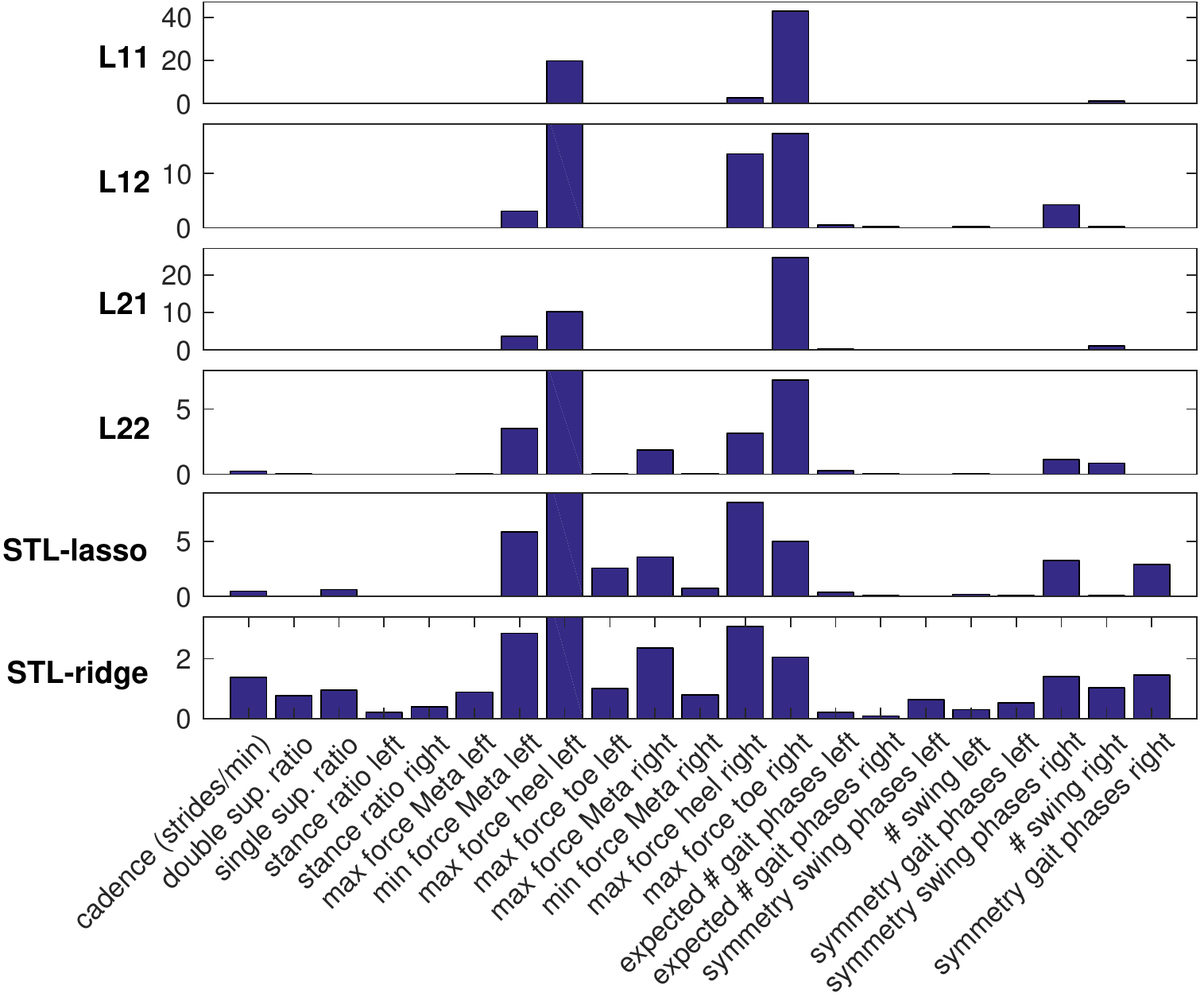}
	\caption{Absolute value of task parameter vector $\vect \alpha_t$ in the Stroke vs. healthy gait classification task.}
\label{fig:stvsh}
\end{figure}

\begin{figure}
	\centering
	\includegraphics[width=\columnwidth]{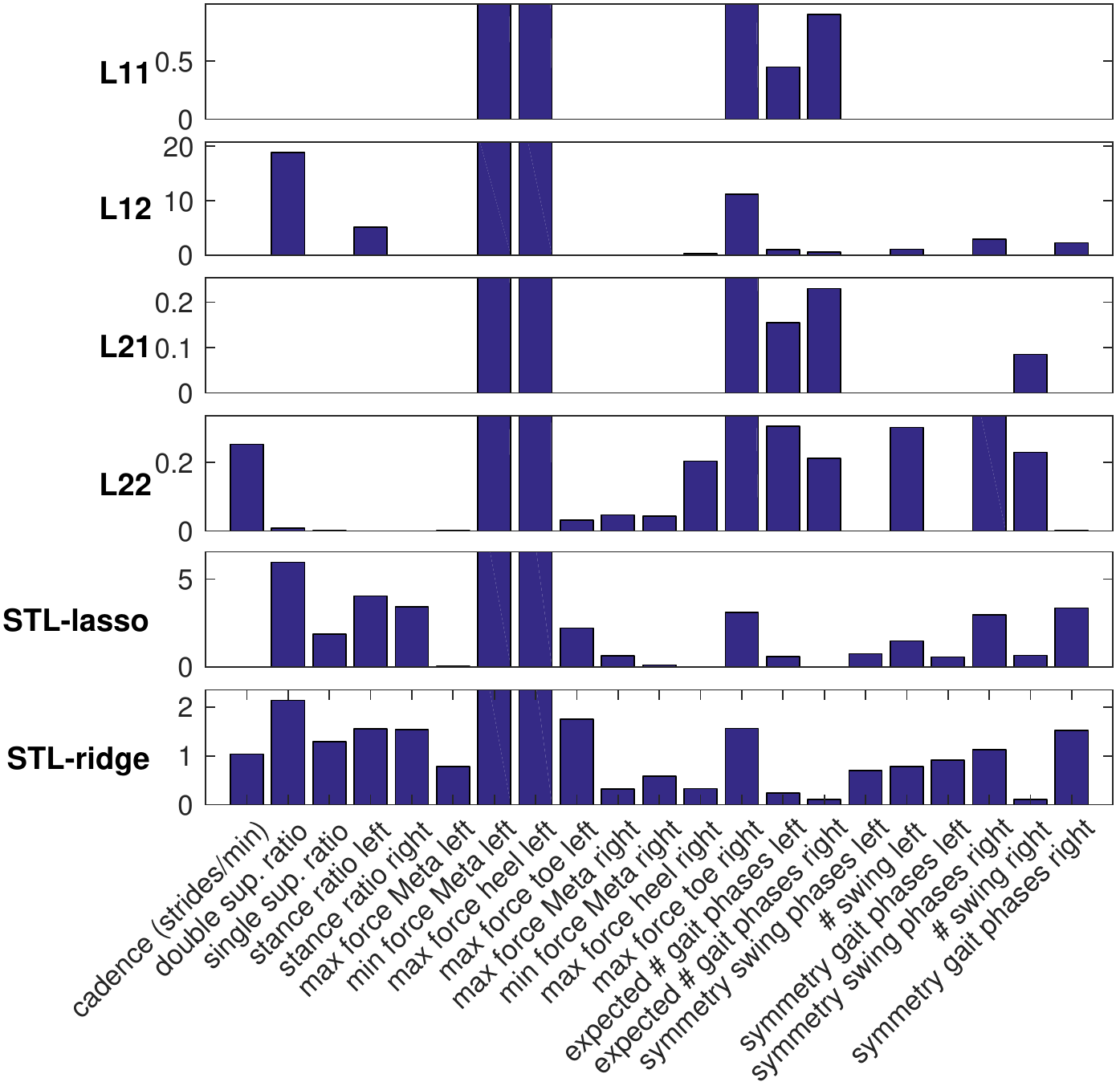}
	\caption{Absolute value of task parameter vector $\vect \alpha_t$ in the Stroke vs. PD gait classification task.}
	\label{fig:stvspd}
\end{figure}

Important gait features identified from gait disorder classification may help better understand the key characteristics that distinguish abnormal gait patterns among different gait disorders and healthy gait. They may also help the target design of treatment and evaluation of rehabilitative progress. In this subsection we present the important gait features that were identified by the used methods in our experiments, for each of the three classification tasks that were evaluated in subsection \ref{sec:classifEval}.
With the important gait features we can understand which of the proposed gait features are more important to classify GCF data from post-stroke or PD patients and healthy subjects.
As described in section~\ref{sec:mtl} for the MMTFL methods, we have $\vect \alpha_t=\vect c \odot \vect \beta_t$.
Vector $\vect \alpha_t$ is the vector of model parameters for task $t$, $\vect c$ vector is used across all tasks, indicating if a feature is useful for any of the tasks, and vector $\vect \beta_t$ is only for task $t$.
In Fig.~\ref{fig:featuresSelected} we plot all vectors $\vect c$ for each MMTFL model as progress bars to show the importance of each feature. In Figures~\ref{fig:pdvsh},~\ref{fig:stvsh} and~\ref{fig:stvspd} we plot the absolute values of the learned task parameter vectors $\vect \alpha_t$ for each MMTFL and STL method for each of the three classification tasks.

Based on the general characteristics of Hemiplegic gait, most commonly seen in stroke, and Parkinsonian gait \cite{BOOK:delisa1998gait} we have the following observations:
\begin{itemize}
	\item The most important feautre is the maximum force at the right toe and the second most selected feature is maximum force at the left heel. These two are strength indicators during toe off and heel strike gait phases. Patients with neurological related diseases, like stroke and PD, may experience weak muscle strength~\cite{BOOK:delisa1998gait}. Circumduction of the affected leg in stroke can also produce different toe contact force signatures. Additionally, slow walking (Bradykinesia) which is characteristic of both stroke and PD gait can have reduced force levels at the toe during push-off~\cite{BOOK:delisa1998gait,JRNL:wu2010statistical}.
	\item Minimum force difference between medial and lateral sides of the metatarsophalangeal joints at the forefoot (see Sec. \ref{sec:balancestrength}) at the left foot is another important feature, which is an indicator of balance. Rigidity, meaning stiff or inflexible muscles, is one of the main symptoms of PD, alongside tremor and slowness of movement. There is usually little or no arm swing to help in balancing the individual~\cite{BOOK:delisa1998gait}. PD patients usually have reduced balance and the algorithm has identified this as an important feature.
	\item Cadence and double support ratio are mobility gait parameters and they are also important in distinguishing healthy vs pathological gait. As discussed before, a common characteristic of stroke and PD subjects is bradykinesia. This in turn affects the double support ratio.
	\item Symmetry of swing phases is found to be another important factor to distinguish pathological gaits for some models. As discussed before, this parameter captures how evenly the swing gait phases are represented in the subject's gait. Circumduction of the affected leg can introduce additional gait phases and thus uneven representation of the detected swing phases.
\end{itemize}
All the rest features are not important and discarded by most of the models, except {MMTFL\{2,2\}, which shows reduced sparsity. These findings are consistent with the literature about the characteristics of PD and stroke patient's gait~\cite{BOOK:delisa1998gait}.

\section{Conclusion}
\label{sec:conclusion}
In this work, we presented the design of an integrative framework for gait disorder diagnosis and advance smart gait rehabilitation. Gait features were developed for different categories including gait phases, mobility, balance and strength. MTFL, an advanced classification method, was used to train the different classification tasks that can classify subject's gait. Data from PD and post-stroke patients, along with healthy subjects were used to evaluate the proposed methods.

The proposed gait features successfully captured the underlying properties of each disease. MTFL was able to construct accurate classifiers based on the given gait parameters to distinguish abnormal gaits. Also it selected the most important gait parameters for this classification task, ignoring the rest. Selected features captured the characteristics of each disease as described in the literature. This study demonstrated the potential to automate
gait analysis of multiple common gait disorders which can benefit the medical
professionals and patients with improved and targeted treatment plans for rehabilitation.

As future work, we intend to provide more comprehensive gait disorder diagnostic tools for more complex gait disorders that are difficult for the clinicians to detect. We plan to assist their assessment process in the clinic, evaluate these analytic systems with properly designed clinical studies, and design new methods for rehabilitation progress evaluation and treatment plan development.

\section{Acknowledgment}
The authors want to acknowledge Dr. Nancy Byl and Ms. Sophia Coo at UCSF for organizing the human subject study,
and the patients who participated in the study for their cooperation.
The authors would like to also thank Dr. Masayoshi Tomizuka at UC Berkeley for his help developing the smart shoes.

\section*{\refname}
\bibliography{IEEEabrv,WL_ACC14}

\begin{thebibliography}{10}
\expandafter\ifx\csname url\endcsname\relax
  \def\url#1{\texttt{#1}}\fi
\expandafter\ifx\csname urlprefix\endcsname\relax\def\urlprefix{URL }\fi
\expandafter\ifx\csname href\endcsname\relax
  \def\href#1#2{#2} \def\path#1{#1}\fi

\bibitem{JRNL:AgingReport}
J.~M. Ortman, V.~A. Velkoff, H.~Hogan, An aging nation: the older population in
  the united states, Washington, DC: US Census Bureau (2014) 25--1140.

\bibitem{MISC:alz}
Alzheimer's facts and figures,
  \url{http://www.alz.org/alzheimers_disease_facts_and_figures.asp}.

\bibitem{MISC:stroke}
Stroke statistics,
  \url{http://www.strokecenter.org/patients/about-stroke/stroke-statistics/}.

\bibitem{MISC:pd}
Statistics on {P}arkinson's, \url{http://www.pdf.org/en/parkinson_statistics}.

\bibitem{BOOK:delisa1998gait}
J.~A. DeLisa, Gait analysis in the science of rehabilitation, Vol.~2, Diane
  Publishing, 1998.

\bibitem{snijders2007neurological}
A.~H. Snijders, B.~P. Van De~Warrenburg, N.~Giladi, B.~R. Bloem, Neurological
  gait disorders in elderly people: clinical approach and classification, The
  Lancet Neurology 6~(1) (2007) 63--74.

\bibitem{jrnl:IMUrehab}
B.-C. Lee, S.~Chen, K.~Sienko, A wearable device for real-time motion error
  detection and vibrotactile instructional cuing 19~(4) (2011) 374--381.

\bibitem{jrnl:camerarehab}
A.~Schmitz, M.~Ye, R.~Shapiro, R.~Yang, B.~Noehren, Accuracy and repeatability
  of joint angles measured using a single camera markerless motion capture
  system, Journal of biomechanics 47~(2) (2014) 587--591.

\bibitem{JRNL:forceplate}
T.~Liu, Y.~Inoue, K.~Shibata, K.~Shiojima, A mobile force plate and
  three-dimensional motion analysis system for three-dimensional gait
  assessment, {IEEE} Sensors J. 12~(5) (2012) 1461--1467.

\bibitem{jrnl:mitss}
S.~Bamberg, A.~Benbasat, D.~Scarborough, D.~Krebs, J.~Paradiso, Gait analysis
  using a shoe-integrated wireless sensor system, IEEE Transactions on
  Information Technology in Biomedicine 12~(4) (2008) 413--423.

\bibitem{jrnl:EMGrehab}
D.~H. Sutherland, The evolution of clinical gait analysis part l:
  kinesiological {E}{M}{G}, Gait \& posture 14~(1) (2001) 61--70.

\bibitem{JRNL:gwin2010removal}
J.~T. Gwin, K.~Gramann, S.~Makeig, D.~P. Ferris, Removal of movement artifact
  from high-density eeg recorded during walking and running, Journal of
  neurophysiology 103~(6) (2010) 3526--3534.

\bibitem{JRNL:sburlea2015detecting}
A.~I. Sburlea, L.~Montesano, R.~C. de~la Cuerda, I.~M.~A. Diego, J.~C.
  Miangolarra-Page, J.~Minguez, Detecting intention to walk in stroke patients
  from pre-movement eeg correlates, Journal of neuroengineering and
  rehabilitation 12~(1) (2015) 1.

\bibitem{JRNL:sant2011new}
A.~Sant'Anna, A.~Salarian, N.~Wickstrom, A new measure of movement symmetry in
  early parkinson's disease patients using symbolic processing of inertial
  sensor data, IEEE Transactions on biomedical engineering 58~(7) (2011)
  2127--2135.

\bibitem{JRNL:tahir2012parkinson}
N.~M. Tahir, H.~H. Manap, Parkinson disease gait classification based on
  machine learning approach, Journal of Applied Sciences 12~(2) (2012) 180.

\bibitem{JRNL:lee2012parkinson}
S.-H. Lee, J.~S. Lim, Parkinson's disease classification using gait
  characteristics and wavelet-based feature extraction, Expert Systems with
  Applications 39~(8) (2012) 7338--7344.

\bibitem{mulroy2003use}
S.~Mulroy, J.~Gronley, W.~Weiss, C.~Newsam, J.~Perry, Use of cluster analysis
  for gait pattern classification of patients in the early and late recovery
  phases following stroke, Gait \& posture 18~(1) (2003) 114--125.

\bibitem{kinsella2008gait}
S.~Kinsella, K.~Moran, Gait pattern categorization of stroke participants with
  equinus deformity of the foot, Gait \& posture 27~(1) (2008) 144--151.

\bibitem{JRNL:chen2005gait}
G.~Chen, C.~Patten, D.~H. Kothari, F.~E. Zajac, Gait differences between
  individuals with post-stroke hemiparesis and non-disabled controls at matched
  speeds, Gait \& posture 22~(1) (2005) 51--56.

\bibitem{JRNL:wu2010statistical}
Y.~Wu, S.~Krishnan, Statistical analysis of gait rhythm in patients with
  parkinson's disease, IEEE Transactions on Neural Systems and Rehabilitation
  Engineering 18~(2) (2010) 150--158.

\bibitem{chantraine2016proposition}
F.~Chantraine, P.~Filipetti, C.~Schreiber, A.~Remacle, E.~Kolanowski,
  F.~Moissenet, Proposition of a classification of adult patients with
  hemiparesis in chronic phase, PloS one 11~(6) (2016) e0156726.

\bibitem{ferrante2011biofeedback}
S.~Ferrante, E.~Ambrosini, P.~Ravelli, E.~Guanziroli, F.~Molteni, G.~Ferrigno,
  A.~Pedrocchi, A biofeedback cycling training to improve locomotion: a case
  series study based on gait pattern classification of 153 chronic stroke
  patients, Journal of neuroengineering and rehabilitation 8~(1) (2011) 1.

\bibitem{JMLR:v17:15-234}
X.~Wang, J.~Bi, S.~Yu, J.~Sun, M.~Song,
  \href{http://jmlr.org/papers/v17/15-234.html}{Multiplicative multitask
  feature learning}, Journal of Machine Learning Research 17~(80) (2016) 1--33.
\newline\urlprefix\url{http://jmlr.org/papers/v17/15-234.html}

\bibitem{obozinski2006multi}
G.~Obozinski, B.~Taskar, M.~Jordan, Multi-task feature selection, Statistics
  Department, UC Berkeley, Tech. Rep 2.

\bibitem{sadeghi2000symmetry}
H.~Sadeghi, P.~Allard, F.~Prince, H.~Labelle, Symmetry and limb dominance in
  able-bodied gait: a review, Gait \& posture 12~(1) (2000) 34--45.

\bibitem{patterson2010evaluation}
K.~K. Patterson, W.~H. Gage, D.~Brooks, S.~E. Black, W.~E. McIlroy, Evaluation
  of gait symmetry after stroke: a comparison of current methods and
  recommendations for standardization, Gait \& posture 31~(2) (2010) 241--246.

\bibitem{JRNL:hubble2015wearable}
R.~P. Hubble, G.~A. Naughton, P.~A. Silburn, M.~H. Cole, Wearable sensor use
  for assessing standing balance and walking stability in people with
  parkinson?s disease: a systematic review, PloS one 10~(4) (2015) e0123705.

\bibitem{mizuike2009analysis}
C.~Mizuike, S.~Ohgi, S.~Morita, Analysis of stroke patient walking dynamics
  using a tri-axial accelerometer, Gait \& posture 30~(1) (2009) 60--64.

\bibitem{particleChase2016}
I.~Papavasileiou, W.~Zhang, S.~Han, Real-time data-driven gait phase detection
  using infinite gaussian mixture model and parallel particle filter, in: 2016
  IEEE First International Conference on Connected Health: Applications,
  Systems and Engineering Technologies (CHASE), 2016, pp. 302--311.
\newblock \href {http://dx.doi.org/10.1109/CHASE.2016.25}
  {\path{doi:10.1109/CHASE.2016.25}}.

\bibitem{wenlongJDSMC}
W.~Zhang, M.~Tomizuka, N.~Byl, A wireless human motion monitoring system for
  smart rehabilitation, ASME. J. Dyn. Sys., Meas., Control 138~(11) (2016)
  111004--1--11004--9.

\bibitem{kaczmarczyk2009gait}
K.~Kaczmarczyk, A.~Wit, M.~Krawczyk, J.~Zaborski, Gait classification in
  post-stroke patients using artificial neural networks, Gait \& posture 30~(2)
  (2009) 207--210.

\bibitem{JRNL:smartshoes}
K.~Kong, M.~Tomizuka, A gait monitoring system based on air pressure sensors
  embedded in a shoe, IEEE/ASME Transactions on Mechatronics 14~(3) (2009)
  358--370.

\bibitem{JRNL:gaitpara}
H.~P. Von~Schroeder, R.~D. Coutts, P.~D. Lyden, E.~Billings, V.~L. Nickel, Gait
  parameters following stroke: a practical assessment, J. Rehabilitation
  Research and Development 32 (1995) 25--25.

\bibitem{byl2015clinical}
N.~Byl, W.~Zhang, S.~Coo, M.~Tomizuka, Clinical impact of gait training
  enhanced with visual kinematic biofeedback: Patients with parkinson?s disease
  and patients stable post stroke, Neuropsychologia 79 (2015) 332--343.

\end{thebibliography}
\end{document}